%% file: main.tex
\documentclass{article}

 \usepackage[preprint]{neurips_2025}

\usepackage[utf8]{inputenc}   
\usepackage[T1]{fontenc}      
\usepackage{amsmath}
\usepackage{amsfonts}         
\usepackage{adjustbox}
\usepackage{graphicx}
\usepackage{booktabs}         
\usepackage{nicefrac}         
\usepackage{microtype}        
\usepackage[table]{xcolor}    
\usepackage{fdsymbol}
\usepackage{listings}
\usepackage{algorithm}
\usepackage{algpseudocode}
\usepackage{caption}
\usepackage{url}
\usepackage{xspace}
\usepackage[colorlinks=true, allcolors=blue]{hyperref} 

\usepackage{multirow}

\definecolor{mygreen}{HTML}{DFFFDF}
\definecolor{myyellow}{HTML}{FFF2A6}

\definecolor{llmoutputcolor}{rgb}{0.1,0.5,0.1} 

\lstdefinestyle{llmstyle}{
    frame=single,
    basicstyle=\ttfamily\small\color{llmoutputcolor}, 
    breaklines=true, 
    breakatwhitespace=false,
}

\newcommand\blankfootnote[1]{%
  \let\thefootnote\relax\footnotetext{#1}%
  \let\thefootnote\svthefootnote%
}
\title{ROMA: Recursive Open Meta-Agent Framework for Long-Horizon Multi-Agent Systems}

\definecolor{myblue}{HTML}{8B98ED}
\newcommand{\fasymbol}{\textcolor{myblue}{*}}
\newcommand{\sasymbol}{\textcolor{myblue}{\textbf{$\vardiamondsuit$}}}
\newcommand{\fa}{\textsuperscript{\fasymbol}}
\newcommand{\sa}{\textsuperscript{\sasymbol}}

\newcommand{\roma}{{ROMA}\xspace}
\newcommand{\github}{\raisebox{-1.5pt}{\includegraphics[height=1.05em]{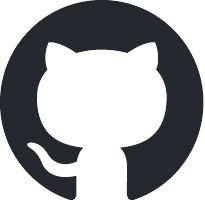}}\xspace}

\author{%
  Salaheddin Alzu'bi$^{1}\fa$ \quad
  Baran Nama$^{1}$ \quad
  Arda Kaz$^{1,3}$ \\
  {\bf Anushri Eswaran}$^{1,4}$ \quad
  {\bf Weiyuan Chen}$^{2}$ \quad
  {\bf Sarvesh Khetan}$^{1,5}$ \quad
  {\bf Rishab Bala}$^{2}$ \quad \\
  {\bf Tu Vu}$^{2}\sa$ \quad {\bf Sewoong Oh}$^{1}\sa$\\
  \\
  $^{1}$Sentient, 
  $^{2}$Virginia Tech, 
  $^{3}$UC Berkeley,
  $^{4}$UC San Diego,
  $^{5}$University of Maryland
}

\begin{document}

\makeatletter
\renewcommand\thanks[1]{%
  \begingroup
  \renewcommand\thefootnote{}%
  \footnotetext{#1}%
  \addtocounter{footnote}{-1}%
  \endgroup
}
\makeatother

\maketitle

\input{sections/abstract}
\input{sections/introduction}
\input{sections/methodology}
\input{sections/experiments}
\input{sections/related_work}
\input{sections/conclusions}
\input{sections/limitations}

\clearpage
\newpage

\bibliographystyle{plain}
\bibliography{neurips_2025}

\clearpage
\newpage

\input{sections/appendices}
\end{document}

%% file: sections/abstract.tex
\begin{center} 
    \github \url{https://github.com/sentient-agi/ROMA}
\end{center}

\begin{abstract}
Current agentic frameworks underperform on long-horizon tasks. As reasoning depth increases, sequential orchestration becomes brittle, context windows impose hard limits that degrade performance, and opaque execution traces make failures difficult to localize or debug. 
We introduce \textbf{ROMA} (\textbf{R}ecursive \textbf{O}pen \textbf{M}eta-\textbf{A}gents), a domain-agnostic framework that addresses these limitations through recursive task decomposition and structured aggregation. \roma decomposes goals into dependency-aware subtask trees that can be executed in parallel, while aggregation compresses and validates intermediate results to control context growth. Our framework standardizes agent construction around four modular roles -- \emph{Atomizer} (which decides whether a task should be decomposed), \emph{Planner}, \emph{Executor}, and \emph{Aggregator} -- which cleanly separate orchestration from model selection and enable transparent, hierarchical execution traces. This design supports heterogeneous multi-agent systems that mix models and tools according to cost, latency, and capability. To adapt \roma to specific tasks without fine-tuning, we further introduce GEPA+, an improved Genetic-Pareto prompt proposer that searches over prompts within ROMA's component hierarchy while preserving interface contracts. We show that \roma, combined with GEPA+, delivers leading system-level performance on reasoning and long-form generation benchmarks. On SEAL-0, which evaluates reasoning over conflicting web evidence, ROMA instantiated with GLM-4.6 improves accuracy by 9.9\% over Kimi-Researcher. On EQ-Bench, a long-form  writing benchmark, ROMA enables DeepSeek-V3 to match the performance of leading closed-source models such as Claude Sonnet 4.5. Our results demonstrate that recursive, modular agent architectures can scale reasoning depth while remaining interpretable, flexible, and model-agnostic.

\blankfootnote{\textsuperscript{\fasymbol}Lead author. \textsuperscript{\sasymbol}Co-senior authors. Correspondence to \texttt{salaheddin@sentient.xyz}.}

\end{abstract}

%% file: sections/introduction.tex
\section{Introduction}

Large language models (LLMs) have enabled rapid progress in \emph{agentic} systems, where multi-step workflows compose model calls, tools, and memory to solve long-horizon tasks such as code generation~\citep{anthropic2025claudecode,cursor2026} and open-domain research~\citep{openai_deep_research,perplexity_deep_research}. These systems now appear in both research prototypes and deployed products, where they often match or exceed strong single-call baselines. Much of this improvement comes from the use of multiple specialized agents that communicate and collaborate to address complex problems~\citep{guo2024large}.

Despite their success, most agentic systems are built in an \emph{ad hoc} manner. Control flow, communication patterns, and context management are typically hard-coded into prompts or orchestration logic, rather than expressed through shared, task-agnostic abstractions. As a result, systems that perform well in one domain rarely transfer cleanly to others. Developing a new agent often requires rebuilding meta-agent structure from scratch, including role definitions, message-passing protocols, and memory interfaces. This fragmentation is reflected in the proliferation of open-source research agents with incompatible designs, while closed-source systems provide little visibility into their internal orchestration.

This lack of standardization leads to recurring failure modes. Agent communication and task decomposition follow no common schema, which makes systems difficult to compare, extend, or reuse. Agent behavior is also opaque. Proprietary systems expose no execution traces, while open-source systems typically provide only unstructured logs, which complicates error localization in deep, branching executions. When failures occur late in a run, there is rarely a principled way to attribute them to earlier planning, retrieval, or aggregation decisions. Additionally, agentic systems suffer from uncontrolled context growth, where accumulating intermediate reasoning, tool outputs, and artifacts degrade performance or exceed context limits.

\input{figures/roma_framework}

In this work, we introduce ROMA (Recursive Open Meta-Agents), a unified framework, inspired by~\cite{xiong-etal-2025-beyond}, that is designed for building multi-agent systems around a prescribed meta-agent structure. ROMA operates over a recursive, node-level control loop that is applied uniformly at every node in a task tree and is composed of four core components: \textbf{Atomizer}, \textbf{Planner}, \textbf{Executors}, and \textbf{Aggregator} (see Figure~\ref{figure:roma_framework}). Starting from a root goal, an \emph{Atomizer} determines whether the current task is atomic. Non-atomic tasks are expanded by a \emph{Planner} into dependency-aware, Mutually Exclusive and Collectively Exhaustive (MECE) subtask graphs. Each resulting subtask independently executes the same control loop: atomic tasks are handled directly by \emph{Executors}, potentially in parallel, while non-atomic tasks are further decomposed. Once all child tasks of a node complete, an \emph{Aggregator} synthesizes, verifies, and compresses their outputs into a higher-level artifact that is returned to the parent. This recursive execution produces a hierarchical trace that mirrors the execution tree.

By standardizing how tasks are decomposed, executed, and aggregated, ROMA replaces ad hoc, task-specific orchestration patterns with a reusable, domain-agnostic abstraction (\texttt{Atomizer} $\rightarrow$ \texttt{Planner} $\rightarrow$ \texttt{Executor} $\rightarrow$ \texttt{Aggregator}) that applies uniformly across domains. Practitioners instantiate new agents by specifying prompts for these components, while ROMA automatically handles control flow, task decomposition, and context propagation. ROMA also improves transparency and traceability of agent behavior. Because the same execution protocol is applied at every node in the task tree, the framework exposes all planning decisions, executions, and aggregations as a structured, hierarchical trace. This makes it possible to inspect and debug failures in deep, branching reasoning processes, where errors are otherwise difficult to localize. Finally, ROMA controls context growth through recursive decomposition and bounded aggregation. Executors operate on localized context, while Aggregators compress intermediate results before passing them upward. Instead of propagating full transcripts, each node returns a concise, verified summary, which reduces context rot~\citep{hong2025context} and enables ROMA to scale to long-horizon tasks while maintaining stable performance.

Beyond addressing these structural limitations, ROMA supports parallel and heterogeneous execution. Independent branches of the task decomposition can execute in parallel, and different models can be assigned to different roles, such as planning, execution, and aggregation. This decouples orchestration from any single foundation model and allows practitioners to exploit heterogeneous model strengths across subtasks while keeping the high-level control loop fixed.

To adapt ROMA to specific tasks without fine-tuning, we further introduce GEPA+, a multi-component prompt optimization method tailored to ROMA's modular architecture. GEPA+ extends prior GEPA-style prompt evolution~\citep{agrawal2025gepa} by jointly optimizing the prompts of the Atomizer, Planner, Executors, and Aggregator. At a high level, GEPA+ works by generating multiple candidate prompt edits in parallel, evaluating them with lightweight judges and verifiers, and merging the best ideas into a single, interface-safe update, which enables efficient exploration without destabilizing the system. This structured, $K$-way proposal and selection process consistently improves downstream performance, yielding 2-6 point absolute accuracy gains while requiring approximately 3–4$\times$ fewer metric evaluations than standard GEPA~\citep{agrawal2025gepa} (see Appendix~\ref{appendix:gepa_vs_gepa_plus}), allowing ROMA to be automatically specialized to new tasks with minimal manual intervention.


We evaluate ROMA across a range of reasoning and long-form generation benchmarks and show that it delivers leading system-level performance. On SEAL-0~\citep{pham2025sealqa}, which tests reasoning over conflicting web evidence, ROMA instantiated with GLM-4.6~\citep{zeng2025glm} improves accuracy by 9.9\% over Kimi-Researcher~\citep{moonshotai2025kimiresearcher}, a reinforcement-learning-tuned deep research agent. On EQ-Bench~\citep{paech2023eq}, a long-form writing benchmark, ROMA enables DeepSeek-V3~\citep{liu2024deepseek} to match the performance of leading closed-source models such as Claude Sonnet 4.5~\cite{anthropic2025introducing}. Taken as a whole, our results demonstrate that recursive, modular agent architectures can scale reasoning depth while remaining interpretable, flexible, and model-agnostic.

To summarize, our main contributions are:
\begin{enumerate}
\item \textbf{ROMA:} We introduce ROMA, a scalable and domain-agnostic meta-agent framework built around a fixed recursive control loop with four roles: Atomizer, Planner, Executors, and Aggregator. ROMA addresses key structural limitations of existing agentic systems, including the lack of a standard schema, opaque execution, and uncontrolled context growth, while allowing practitioners to exploit heterogeneous model strengths across subtasks.

\item \textbf{GEPA+:} We develop GEPA+, a prompt optimization method tailored to ROMA's modular architecture, which jointly optimizes the prompts of individual components through a structured, multi-candidate proposal and selection process. GEPA+ enables efficient task adaptation without fine-tuning and improves performance with minimal manual intervention.

\item \textbf{Leading system-level empirical performance:} We evaluate ROMA across a range of reasoning and long-form generation benchmarks, including SEAL-0~\citep{pham2025sealqa}, FRAMES~\citep{krishna-etal-2025-fact}, SimpleQA~\citep{wei2024measuring}, EQ-Bench~\citep{paech2023eq}, and AbGen~\citep{zhao-etal-2025-abgen}. We show that ROMA instantiated with GLM-4.6 improves accuracy on SEAL-0 by 9.9\% relative to Kimi-Researcher, a reinforcement-learning–tuned deep research agent, and that, when combined with GEPA+, ROMA enables DeepSeek-V3 to match the performance of leading closed-source models, including Claude Sonnet 4.5, on EQ-Bench.
\end{enumerate}

%% file: figures/roma_framework.tex
\begin{figure}[t]
    \centering
    \includegraphics[width=1.0\linewidth]{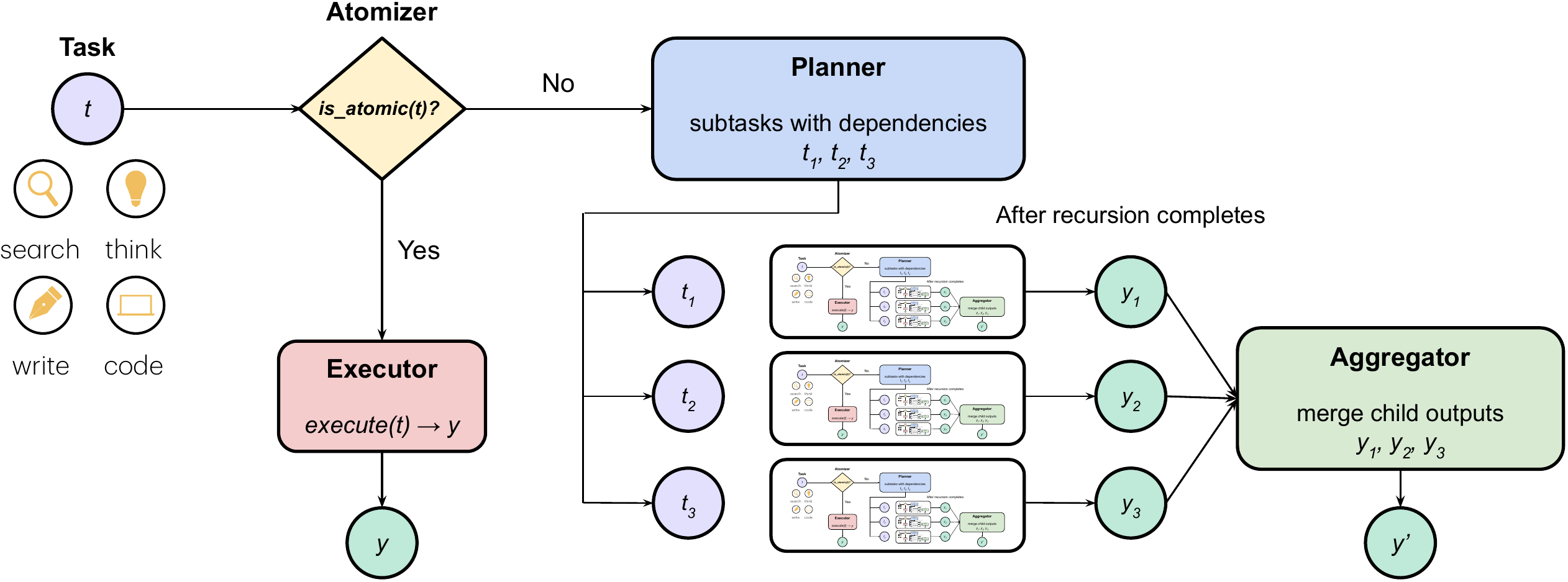}
    \caption{Overview of ROMA's recursive architecture. An \emph{Atomizer} determines whether a task is atomic. Atomic tasks are executed directly, while non-atomic tasks are decomposed into subtasks by a \emph{Planner}. Each subtask is executed recursively by \emph{Executors}, after which an \emph{Aggregator} merges the outputs of all descendants to produce the final result.}
    \label{figure:roma_framework}
    \vspace{-5pt}
\end{figure}

%% file: sections/methodology.tex
\section{Methodology} 
\label{section:method}
In this section, we present the ROMA framework and its execution semantics. Additionally, we introduce GEPA+ for component-wise prompt optimization. 

\vspace{-5pt}
\subsection{ROMA}

ROMA is a recursive meta-agent that solves a task by alternating between \emph{decomposition} and \emph{execution} until all subproblems are atomic, after which evidence is \emph{aggregated} bottom-up to produce the parent's answer (Algorithm~\ref{algorithm:recursive_solver}). For each task or subtask in ROMA's dynamically constructed task graph, a controller invokes the same four components: \textbf{Atomizer}, \textbf{Planner}, \textbf{Executors}, and \textbf{Aggregator} (see Figure~\ref{figure:roma_framework}). This design generalizes heterogeneous recursive planning routines previously used in writing assistants~\citep{xiong-etal-2025-beyond} to a domain-agnostic setting, where composition, retrieval, reasoning, and programmatic manipulation coexist within a single control loop and are scheduled with explicit dependencies.

Let \(t\) denote a task with a textual specification and optional constraints. ROMA exposes four operators:
\[
\begin{aligned}
\texttt{is\_atomic}(t) &\to \{\texttt{True}, \texttt{False}\} & \texttt{(Atomizer)} \\
\texttt{plan}(t) &\to \mathcal{S}(t) & \texttt{(Planner)}  \\
\texttt{execute}(t) &\to y_t & \texttt{(Executor)}  \\
\texttt{aggregate}\!\left(\{y_s\}_{s \in \mathcal{S}(t)}\right) &\to y_{t-1} & \texttt{(Aggregator)} 
\end{aligned}
\]

If \(\texttt{is\_atomic}(t)\) returns \texttt{True}, the Atomizer deems the task \emph{atomic} and the system bypasses planning, invoking the Executor to directly apply $\texttt{execute}(t)$ to produce $y_t$. Otherwise, the Planner applies $\texttt{plan}(t)$ to decompose $t$ into a dependency-aware subtask graph \(\mathcal{S}(t)\), which can be seen as a directed acyclic graph (DAG), where nodes correspond to subtasks and directed edges encode execution dependencies. In the simplest case, all subtasks depend only on the parent task and are independent of one another (e.g., $\mathcal{S}(t)=\{t \rightarrow t_1, t \rightarrow t_2, t \rightarrow t_3\}$). More generally, $\mathcal{S}(t)$ may encode dependencies among subtasks (e.g., $\mathcal{S}(t)=\{t \rightarrow t_1, t \rightarrow t_2, t \rightarrow t_3, t_1 \rightarrow t_3, t_2 \rightarrow t_3\}$, in which $t_3$ is blocked until $t_1$ and $t_2$ complete). The same control loop is then applied independently to each subtask $s \in \mathcal{S}(t)$, and subtasks may execute in parallel when dependencies allow. Once all subtasks complete, the Aggregator combines their outputs through $\texttt{aggregate}\!\left(\{y_s\}_{s \in \mathcal{S}(t)}\right)$ to produce the parent task's result $y_{t-1}$. This plan--execute--aggregate loop is the sole control flow in ROMA, which makes the framework task-agnostic, transparent, and extensible.

\input{figures/roma_hierarchical_exe_flow}

Figure~\ref{figure:roma_roma_hierarchical_exe_flow} illustrates ROMA's hierarchical execution flow. Tasks are decomposed top-down through planning, while results are combined bottom-up through aggregation. Executors operate only on atomic subtasks, producing intermediate outputs that are aggregated into higher-level artifacts. The same recursive control loop is applied at every node in the task tree, reinforcing ROMA's generality across domains and task types. Below, we discuss ROMA's core properties.

\vspace{-5pt}
\paragraph{Top-down decomposition and bottom-up aggregation: } The Planner decomposes a non-atomic task $t$ into a \emph{mutually exclusive}, \emph{collectively exhaustive} (MECE) subtask graph $\mathcal{S}(t)$ that maximizes parallelism and minimizes redundancy. Subtasks are defined at the intent level and connected by explicit dependency edges when left-to-right precedence is required. The result is a compact ordered DAG that supports parallel execution.

A parent's output is not a raw concatenation of child results. Instead, the Aggregator applies $\texttt{aggregate}\!\left(\{y_s\}_{s \in \mathcal{S}(t)}\right)$ to distill and normalize child outputs into the parent's target form (e.g., a paragraph, a table, or a JSON schema). Executors are restricted to operate on local context, while Aggregators perform cross-cutting synthesis and relevance-preserving compression. This design mitigates context explosion and allows the effective working set to exceed a single model's context window.

\vspace{-5pt}
\paragraph{Dependency-aware parallel execution:} At each level of the task tree, Planner returns a dependency-aware subtask graph $\mathcal{S}(t)$ with explicit left-to-right dependency edges that constrain execution. Subtasks with no unmet dependencies may execute in parallel, while dependent subtasks are scheduled as soon as their predecessors complete. This dependency-aware scheduling avoids global barriers and enables latency-efficient execution along the critical path, while preserving correctness when subtasks require inputs from earlier siblings (e.g., \textit{``identify the highest-scoring player'' $\rightarrow$ ``retrieve their age''}).

\vspace{-5pt}
\paragraph{Task types and type-specialized execution: } Each task node is annotated with a \texttt{task\_type} drawn from a small, expressive set:

\begin{enumerate}
    \item \texttt{search} (retrieval): retrieve documents, facts, or tool outputs from external sources.
    \item \texttt{think} (reasoning): synthesize intermediate conclusions (e.g., derivations or chain-of-thought (CoT)).
    \item \texttt{write} (composition): compose structured or expository outputs from evidence (e.g., sections, summaries, answers).
    \item \texttt{code} (programmatic manipulation): write or execute code to transform data or invoke structured tools.
\end{enumerate}

The Atomizer's atomicity decision is orthogonal to \texttt{task\_type}. A node may be non-atomic and handed to the Planner, or atomic and handed to an Executor, independent of its type. This separation allows ROMA to decouple \emph{how a task is decomposed} from \emph{how a leaf computation is carried out}. In our implementation, each \texttt{task\_type} routes to a type-specialized Executor with distinct prompting strategies (e.g., ReAct~\citep{yao2023react}, CodeAct~\citep{wang2024executable}, or CoT~\citep{wei2022chain}) and model choices, which can be selected to trade off cost, latency, and quality (see Figure~\ref{figure:roma_roma_hierarchical_exe_flow}).

\vspace{-5pt}
\paragraph{Modular components and optimization hooks: } ROMA's components are implemented as modular \texttt{DSPy}~\citep{khattab2024dspy} programs with \emph{typed input/output signatures}. This design provides \emph{structured interfaces} between the Atomizer, Planner, Executor, and Aggregator, provides structured interfaces betweenenables composability by allowing modules to be swapped while preserving type compatibility, and exposes built-in optimization hooks for prompts and weights via DSPy optimizers. DSPy's \emph{``programming-not-prompting''} abstractions allow these components to be declared as executable modules and compiled into high-performing LM invocations, which we find essential for building and optimizing a multi-component agent system.

\vspace{-5pt}
\paragraph{Artifacts, tools, and safe execution: } ROMA is designed for tool-rich and high-throughput workloads. Intermediate artifacts (e.g., plans, notes, citations, and datasets) are persisted to an object store and exposed to downstream nodes through typed module signatures, rather than embedded directly in prompts. This allows artifacts to be reused across the task tree without inflating prompt context. Programmatic subtasks are handled by the \texttt{code} task type, which uses a \emph{sandboxed runtime} to execute user- or model-generated code in a safe, isolated environment and to interact with external tools through standard interfaces, such as the Model Context Protocol (MCP)~\citep{anthropic2024introducing}. This design supports efficient multi-agent execution while preserving isolation, observability, and safety.

\captionsetup[algorithm]{position=bottom} 

\begin{algorithm}[t]
\begin{algorithmic}[1]
\Procedure{Solve}{task}
    \If{is\_atomic(task)} \Comment{Step 1: Atomizer}
        \State \Return execute(task) \Comment{Step 2: Executor}
    \Else
        \State subtasks $\gets$ plan(task) \Comment{Step 2: Planner}
        \State results $\gets$ [ ] \Comment{Initialize empty list}
        \ForAll{subtask $\in$ subtasks}
            \State results.append(\textsc{Solve}(subtask)) \Comment{Recursive call}
        \EndFor
        \State \Return aggregate(results) \Comment{Step 3: Aggregator}
    \EndIf
\EndProcedure
\end{algorithmic}
\caption{\textbf{ROMA:} A recursive control loop with \textit{Atomizer}, \textit{Planner}, and \textit{Aggregator}.}
\label{algorithm:recursive_solver}
\end{algorithm}

\paragraph{Putting it together -- the ROMA control loop:}
Algorithm~\ref{algorithm:recursive_solver} summarizes ROMA's recursive control loop. Information flows \emph{top-down} during planning, \emph{left-to-right} within each level to respect dependency constraints, and \emph{bottom-up} during aggregation. Executors are invoked only on atomic tasks, while Aggregators return \emph{parent-scoped} results rather than raw child outputs. This structured separation of planning, execution, and aggregation allows ROMA to scale to long-horizon tasks while keeping each component focused, interpretable, and easy to optimize.

\vspace{-5pt}
\subsection{Multi-component prompt optimization with GEPA+}
In this section, we introduce GEPA+, a multi-proposer extension of DSPy's GEPA~\citep{agrawal2025gepa} that is tailored to ROMA's modular, multi-component architecture. Whereas classical GEPA uses a \emph{single} reflection model to propose an instruction/prompt update for one module given execution traces and feedback, GEPA+ replaces this with a $K-$\textit{way proposer} that (i) \emph{generates} diverse prompt edits in parallel, (ii) \emph{re-ranks} them using learned and verifier-based signals, and (iii) \emph{merges} the strongest candidates into a single, interface-safe update. The result is a minimal-change optimizer that more effectively explores the local edit space while preserving cross-module contracts. Below, we detail the optimization procedure. More implementation details can be found in our code.\footnote{\url{https://github.com/sentient-agi/gepa-plus}.}

\vspace{-5pt}
\paragraph{Problem setup: } Let the system consist of modules \(m \in \mathcal{M}\) (Atomizer, Planner, type-specific Executors, Aggregator), each with a current instruction \(I_m\). Given a development set \(\mathcal{D}\) with automatic checks, GEPA+ searches for \emph{delta} edits \(\Delta I_m\) that improve a utility function \(U(\{I_m\}_{m\in\mathcal{M}};\mathcal{D})\) under budget and stability constraints, while maintaining module interface invariants (e.g., required fields in \texttt{search} outputs).

\vspace{-5pt}
\paragraph{Diverse proposals: } For a target module \(m\), GEPA+ samples \(k\) \emph{independent} proposal candidates \(\Pi_m=\{\pi_1,\ldots,\pi_k\}\) in parallel, conditioned on execution traces and feedback. Diversity is induced through a mixture of base LLMs, decoding settings (e.g., with different temperatures and random seeds), and optional reflection prompts. Proposals are module-aware, for example, Planner edits must preserve MECE and dependency constraints, while Aggregator edits must respect expected input schemas from sibling modules.

\vspace{-5pt}
\paragraph{Reranking with judges and verifiers: } Each candidate \(\pi_i\) is scored by a composite function
\vspace{-2pt}
\[ \mathcal{J}(\pi_i \mid \mathcal{D}) = \alpha\, \text{Judge}(\pi_i) + \beta\, \text{Verifier}(\pi_i;\mathcal{D}) - \gamma\, \text{ContractViolations}(\pi_i), \]
where \textit{Judges} (i.e., LLM-as-a-Judge~\citep{zheng2023judging}) assign rubric-based quality scores on \textit{held-out} execution traces; \textit{Verifiers} run fast, task-specific checks (e.g., unit tests for \texttt{code}, citation and consistency checks for \texttt{search} and \texttt{write}, and interface conformance checks for the Aggregator), and \textit{ContractViolations} penalize edits that break typed I/O or ROMA DAG invariants (i.e., acyclic, dependency-consistent task graphs). The top-\(n\) candidates \(\Pi_m^{(n)}\) are selected according to \(\mathcal{J}(\cdot)\).

\vspace{-5pt}
\paragraph{Structured merge of candidate edits: } GEPA+ consolidates the selected candidates \(\Pi_m^{(n)}\) into a single update \(\widehat{\Delta I_m}\) via a contract-preserving \emph{merger} that performs: (i) \textbf{Atomization}, which decomposes each proposal into atomic edits (e.g., adding one constraint, replacing one example, or rephrasing one instruction); \textbf{Conflict} resolution, which resolves incompatible edits by retaining higher-scoring variants while demoting alternatives to auxiliary examples or comments; and (iii) \textbf{Deduplication and alignment}, which clusters near-duplicate edits and normalizes terminology to preserve module contracts and MECE constraints. The merger guarantees schema compliance for module 
$m$ and compatibility with neighboring modules.

\vspace{-5pt}
\paragraph{Budget and stability: } Parallel proposal generation reduces wall-clock latency while respecting fixed token and time budgets. We cap $k$, $n$, judge calls, and verifier runs per round. GEPA+ favors minimal edits through delta-size penalties, which helps preserve trace stability and avoids prompt drift.

\vspace{-5pt}
\paragraph{Summary:} Compared to single-proposer GEPA, GEPA+ widens local search through parallel, diverse proposals, filters candidates using task-aware judges and verifiers, and fuses improvements via a contract-preserving merger. In our experiments, this \textit{``propose$\times k \rightarrow$ rerank $\rightarrow$ merge''} loop consistently yields higher utility under the same budget, while keeping ROMA's module interfaces intact. See Appendix~\ref{appendix:gepa_vs_gepa_plus} for a detailed comparison between GEPA and GEPA+.

%% file: figures/roma_hierarchical_exe_flow.tex
\begin{figure}[t]
    \centering
    \includegraphics[width=0.8\linewidth]{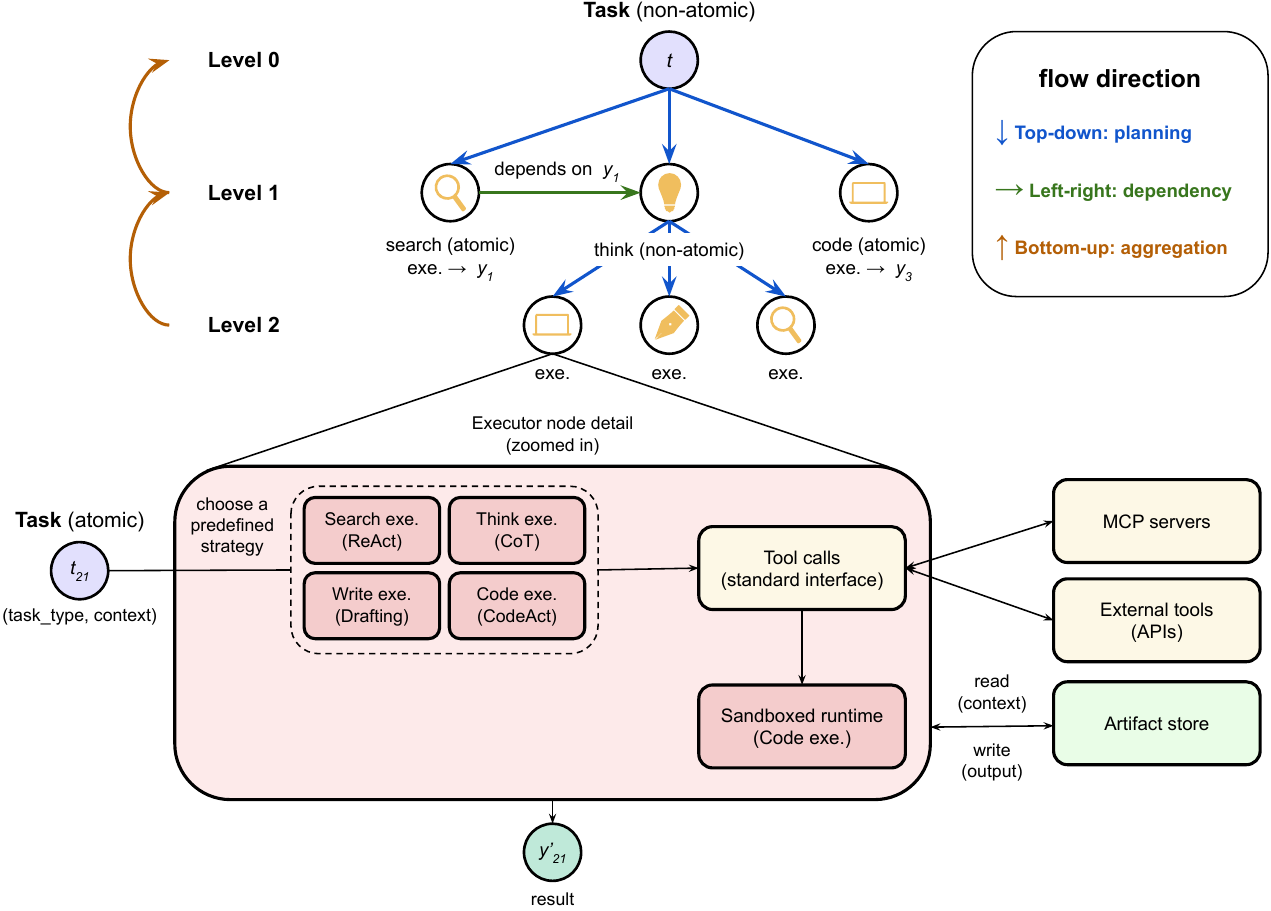}
    \caption{ROMA's hierarchical execution flow. Non-atomic tasks are decomposed top-down through planning, with left-to-right dependencies guiding execution, while results are combined bottom-up through aggregation. Executors operate on atomic subtasks, producing intermediate outputs that are aggregated into higher-level artifacts; the inset shows a zoomed-in executor node with tool interaction.}
    \label{figure:roma_roma_hierarchical_exe_flow}
    \vspace{-5pt}
\end{figure}

%% file: sections/experiments.tex
\section{Experiments} 
\vspace{-5pt}
\subsection{Experimental Setup}
We evaluate ROMA on a broad suite of benchmarks that target complementary capabilities. Specifically, we evaluate on SEAL-0~\citep{pham2025sealqa}, which tests reasoning over conflicting web evidence; FRAMES~\citep{krishna-etal-2025-fact} and SimpleQA~\citep{wei2024measuring}, which evaluate multi-hop factual reasoning and question answering, respectively; EQ-Bench~\citep{paech2023eq}, which focuses on long-form writing, and AbGen~\citep{zhao-etal-2025-abgen}, which assesses a model's ability to design rigorous ablation studies. 

Though ROMA is model-agnostic and supports heterogeneous model assignment across roles, our experiments primarily instantiate ROMA with a single base model per task, which allows us to isolate the framework's contribution without introducing additional variability from cross-model interactions. Specifically, we use GLM~\citep{zeng2025glm} for search- and reasoning-intensive benchmarks (SEAL-0, FRAMES, and SimpleQA) and DeepSeek\citep{liu2024deepseek} for long-form generation benchmarks (EQ-Bench and AbGen). We leave a systematic exploration of heterogeneous model assignments across roles to future work. Additionally, we rely on search results provided by the GPT-5-mini search interface for search-intensive tasks. ROMA operates over the returned evidence using its recursive decomposition and aggregation mechanisms. See Appendix~\ref{appendix:config_test} for complete experimental configurations.

We do not use a fixed set of baselines across all experiments; instead, most baselines are taken directly from the corresponding benchmark or chosen per benchmark to reflect the most relevant comparisons. For example, on SEAL-0 we compare against Kimi-Researcher~\citep{moonshotai2025kimiresearcher}, a reinforcement-learning-tuned autonomous agent specialized for search and reasoning, and Perplexity Deep Research~\citep{perplexity_deep_research}, a closed-source system designed for web search-centric research workflows. This task-specific baseline selection avoids misleading comparisons between systems optimized for different goals.

We apply and report GEPA/GEPA+ only on EQ-Bench, where prompt optimization has a material impact on long-form writing performance; all other experiments use fixed, hand-written component prompts to isolate the contribution of ROMA's execution structure. In our implementation, GEPA+ uses a multi-proposer setup in which candidate prompt edits are generated in parallel by a small set of diverse models, including GPT-5~\citep{singh2025openai}, Claude Sonnet 4.5~\cite{anthropic2025introducing}, and Gemini 2.5 Flash~\citep{comanici2025gemini}, with different decoding temperatures. Candidate proposals are evaluated by a judge model (Claude Sonnet 4.5) and merged by a dedicated merger model (GPT-5) into a single interface-safe update, with the top-$n$ ($n=2$) proposals retained for merging. This design balances proposal diversity with stability and allows GEPA+ to explore the local prompt-edit space efficiently. See Appendix~\ref{appendix:gepa-plus-config} for GEPA+ configuration details.

Finally, we use GPT-4o-mini~\citep{hello_gpt4o} as an LLM judge to evaluate model outputs on SEAL-0, FRAMES, and SimpleQA, following the standard evaluation procedures of these benchmarks (see Appendix~\ref{appendix:judging_prompt} for the full judging prompt). For EQ-Bench, evaluation is performed using Claude Sonnet 4~\citep{introducing_claude4}, consistent with the benchmark's official long-form writing protocol. For AbGen, we follow the benchmark's original evaluation setup and use GPT-4.1-mini~\citep{introducing_gpt41} as the judge to score ablation study designs along multiple Likert-scale dimensions. Across all benchmarks, we adhere to the respective published evaluation protocols to ensure fair and comparable results.

\subsection{Results and discussion}
\input{tables/sealqa}
\paragraph{SEAL-0 (reasoning over conflicting web evidence):} Table~\ref{table:seal0_results} shows our results on SEAL-0, a benchmark of 111 questions designed to evaluate reasoning over noisy and conflicting web evidence. In addition to reported results from~\cite{pham2025sealqa}, we run three baselines ourselves, including Perplexity Sonar Reasoning Pro~\citep{perplexity_sonar_pro}, Perplexity Deep Research~\citep{perplexity_deep_research}, and Open Deep Search~\citep{alzubi2025open}, using their default or recommended hyperparameters.\footnote{For Open Deep Search, we used Serper.dev as the search provider, LiteLLM with DeepSeek-R1 as the reasoning model, self-hosted Infinity Embeddings for reranking, and SmolAgents for enhanced reasoning. See \url{https://github.com/sentient-agi/OpenDeepSearch} for more details.} ROMA achieves the best overall performance with 45.9\% accuracy, outperforming both open-source and closed-source baselines. This corresponds to a 9.9\% absolute improvement over Kimi-Researcher (36.0\%), the strongest prior open research agent, and a 14.4\% absolute improvement over Perplexity Deep Research (31.5\%), the best closed-source system evaluated. ROMA also substantially outperforms its underlying base model (GLM-4.6, 14.5\%), indicating that the gains stem from the agent architecture rather than model scale alone. 

ROMA's gains indicate that how evidence is decomposed and reconciled is critical. ROMA's recursive decomposition isolates conflicting evidence into focused subtasks, while structured aggregation explicitly compares and synthesizes results, reducing error accumulation from reasoning over many documents at once. This suggests that dependency-aware decomposition and aggregation are particularly effective for search-intensive reasoning tasks.

\input{tables/frames}
\paragraph{FRAMES (multi-hop reasoning):} Table~\ref{table:frames_results} reports results on FRAMES, which consists of 824 multi-hop factual questions requiring integration of information across multiple Wikipedia pages. ROMA achieves the highest accuracy on this benchmark at 82.3\%, exceeding the performance of both open- and closed-source baselines. Kimi-Researcher ranks second at 78.8\%, whereas general-purpose reasoning models such as DeepSeek-R1 and Llama-3.1-70B lag far behind, remaining below 35\% accuracy.

In contrast to SEAL-0, FRAMES primarily emphasizes compositional reasoning across ordered hops. ROMA improves over search-augmented baselines by maintaining coherence across successive reasoning steps. ROMA's execution model supports this by treating each hop as a separate reasoning unit and combining intermediate results only after all required sub-questions are resolved, which helps limit error accumulation in longer reasoning chains.

\input{tables/simpleqa}
\paragraph{SimpleQA (factual question answering): } Table~\ref{table:simpleqa_results} reports results on SimpleQA, which consists of 4,326 short-form factual questions with a single unambiguous answer, spanning a wide range of topics. Overall performance on SimpleQA is high across both open-source and closed-source systems, reflecting the relatively direct nature of the task. The strongest closed-source system, Liner Pro Reasoning, achieves 95.3\% accuracy, while ROMA reaches 93.9\%, ranking as the best-performing open-source system and narrowly trailing the top closed-source result. Kimi-Researcher follows closely at 93.6\%.

SimpleQA primarily stresses precise retrieval and factual verification, rather than deep multi-step reasoning. As a result, systems with effective search capabilities perform well even without explicit task decomposition. Nevertheless, ROMA remains competitive by reliably routing SimpleQA queries to focused retrieval-oriented execution, demonstrating that its general-purpose agent architecture does not incur overhead on simpler tasks. Overall, SimpleQA shows that ROMA maintains strong performance on straightforward factual queries, narrowing the gap between open- and closed-source systems in practical web-based question answering.

\input{tables/eqbench}
\paragraph{EQ-Bench (long-form writing):} EQ-Bench evaluates long-form writing, emphasizing a model's ability to plan, maintain coherence, and sustain narrative structure across multiple turns. Each task is executed over eight turns of approximately 1,000 words each, and outputs are scored for narrative quality, character depth, and structural consistency.

Using ROMA instantiated with DeepSeek-V3 and default prompts,\footnote{We follow the configuration described in Appendix B, except that the default Planner is replaced with DeepSeek-V3.} we obtain a long-form writing score of 71.9\%. Applying GEPA+ prompt optimization yields a substantial improvement to 79.8\%, which matches the performance of leading closed-source models such as Claude Sonnet 4.5 (see Table~\ref{table:eqbench_longform_writing}). Here, our reported results reflect the best performance across five independent runs. We note that evaluation is performed using Claude Sonnet 4, which may introduce bias in favor of Claude-family models. This improvement indicates that prompt design plays a critical role for long-horizon generation tasks. Qualitative analysis indicates that GEPA+ strengthens both high-level planning and execution guidance (see Appendices~\ref{appendix:gepa-plus-prompt-atomizer},~\ref{appendix:gepa-plus-prompt-planner}, and~\ref{appendix:gepa-plus-prompt-executor} for full prompts). Optimized Planner prompts emphasize explicit story structure and pacing, while optimized Executor prompts encourage concrete scene construction through actions, dialogue, and sensory detail, leading to more coherent character arcs and more consistent narrative flow throughout the narrative. Overall, these results suggest that ROMA's modular execution framework provides a strong foundation for long-form generation, with GEPA+ enabling targeted improvements without altering the underlying control structure. Table~\ref{table:roma_cost_breakdown} shows the average per-chapter cost, token usage, and end-to-end latency for ROMA instantiated with DeepSeek-V3.1. Overall, ROMA achieves competitive quality with manageable cost and latency, demonstrating that structured multi-component execution can support long-form generation without prohibitive computational overhead.
\input{tables/eqbench_cost}

\input{tables/abgen}
\paragraph{AbGen (designing ablation studies for scientific research):} AbGen evaluates an agent's ability to design open-ended ablation studies given a concrete research context, targeting expert-level scientific reasoning. We evaluate ROMA on a randomly sampled set of 100 AbGen questions, where each task provides background material from a research paper (including introduction, methodology, and main experiment sections) and prompts the agent to propose an ablation study design. Performance is assessed using an LLM-as-a-Judge \citep{vu2024foundational} protocol with GPT-4.1-mini~\citep{introducing_gpt41}, following the AbGen evaluation scheme. Outputs are scored on Importance, Faithfulness, Soundness, and Overall quality using a Likert scale from 1 to 5, referencing ground-truth ablations extracted from the original papers, and results are averaged across all samples.

Our results are shown in Table~\ref{table:abgen_results}. ROMA achieves the strongest overall performance among evaluated systems, leading on Importance, Soundness, and Overall metrics compared to state-of-the-art single-model baselines. These results suggest that ROMA generalizes beyond factual reasoning to high-stakes, domain-specific scientific tasks, where structuring complex experimental considerations is essential. ROMA's performance on Faithfulness is comparatively lower (ranking 4th among 20 settings), which we attribute to slight degradation in alignment with the original research context as inference depth and computation increase. Importantly, this does not substantially affect the overall quality of the generated ablation designs, indicating that ROMA's hierarchical organization improves high-level experimental reasoning even when fine-grained contextual alignment is imperfect.

%% file: tables/sealqa.tex
\begin{table}[h]
\centering
\caption{SEAL-0 accuracy for closed-source and open-source systems. $^\star$ denotes results taken from the SEAL-0 leaderboard. $^\diamond$ denotes baseline evaluations run by the authors. $^\spadesuit$ denotes systems that were not publicly available at the time of evaluation.}
\label{table:seal0_results}
\vspace{5pt}
\begin{tabular}{lcc}
\toprule
\textbf{Model} & \textbf{Search} & \textbf{SEAL-0 (\%)} \\
\midrule
\rowcolor{myyellow}\multicolumn{3}{c}{\textit{Closed‑source}} \\ \addlinespace
Grok-3$^\star$ & $\checkmark$ & 5.4 \\
Gemini 2.5 Flash$^\star$ & $\checkmark$ & 13.5 \\
Perplexity Sonar Reasoning Pro$^\diamond$ & $\checkmark$ & 13.5 \\
o3$^\star$ & $\checkmark$ & 15.3 \\
o3-pro$^\star$ & $\checkmark$ & 18.9 \\
Gemini 2.5 Pro$^\star$ & $\checkmark$ & 19.8 \\
Grok-4$^\star$ & $\checkmark$ & 20.7 \\
Perplexity Deep Research$^\diamond$ & $\checkmark$ & 31.5 \\
\midrule
\rowcolor{mygreen}\multicolumn{3}{c}{\textit{Open‑source}} \\ \addlinespace
DeepSeek-R1$^\star$ & $\checkmark$ & 4.5 \\
Qwen3-235B-A22B$^\star$ & $\checkmark$ & 5.4 \\
Open Deep Search (DeepSeek-R1)$^\diamond$ & $\checkmark$ & 9.9 \\
GLM-4.6 & $\checkmark$ & 14.5 \\
Kimi-Researcher$^\spadesuit$~\citep{moonshotai2025kimiresearcher} & $\checkmark$ & 36.0 \\
ROMA (GLM-4.6) & $\checkmark$ & \textbf{45.9} \\
\bottomrule
\end{tabular}
\end{table}
\vspace{-5pt}

%% file: tables/frames.tex
\begin{table}[h]
\centering
\caption{FRAMES accuracy for closed-source and open-source systems. $^\star$ denotes results taken from~\citep{alzubi2025open}. $^\spadesuit$ denotes systems that were not publicly available at the time of evaluation.}
\label{table:frames_results}
\vspace{5pt}
\begin{tabular}{lcc}
\toprule
\textbf{Model} & \textbf{Search} & \textbf{FRAMES (\%)} \\
\midrule
\rowcolor{myyellow}\multicolumn{3}{c}{\textit{Closed‑source}} \\ \addlinespace
Perplexity$^\star$ & $\checkmark$ & 42.4 \\
Perplexity Sonar Reasoning Pro$^\star$ & $\checkmark$ & 44.4 \\
GPT-4o$^\star$ & $\times$ & 50.5 \\
GPT-4o Search Preview$^\star$ & $\checkmark$ & 65.6 \\
\midrule
\rowcolor{mygreen}\multicolumn{3}{c}{\textit{Open‑source}} \\ \addlinespace
DeepSeek-R1$^\star$ & $\times$ & 30.1 \\
Llama-3.1-70B$^\star$ & $\times$ & 34.3 \\
GLM-4.6 & $\checkmark$ & 71.2 \\
Open Deep Search (DeepSeek-R1) \citep{alzubi2025opendeepsearch2}$^\star$ & $\checkmark$ & 75.3 \\
Kimi-Researcher$^\spadesuit$~\citep{moonshotai2025kimiresearcher} & $\checkmark$ & 78.8 \\
ROMA (GLM-4.6) & $\checkmark$ & \textbf{82.3} \\

\bottomrule
\end{tabular}
\end{table}
\vspace{-5pt}

%% file: tables/simpleqa.tex
\begin{table}[t]
\centering
\caption{ROMA achieves the highest accuracy among open-source systems on SimpleQA. $^\star$ denotes results taken from the SimpleQA leaderboard.}
\vspace{5pt}
\label{table:simpleqa_results}
\begin{tabular}{lcc}
\toprule
\textbf{Model} & \textbf{Search} & \textbf{SimpleQA (\%)} \\
\midrule
\rowcolor{myyellow}\multicolumn{3}{c}{\textit{Closed‑source}} \\ \addlinespace
o1-mini$^\star$ & $\times$ & 7.6 \\
GPT-4o-mini-2024-07-18$^\star$ & $\times$ & 9.5 \\
o3-mini-low$^\star$ & $\times$ & 13.0 \\
o3-mini$^\star$ & $\times$ & 13.4 \\
o3-mini-high$^\star$ & $\times$ & 13.8 \\
Grok 3 mini Beta~\citep{grok3_beta} & $\times$ & 21.7 \\
Claude 3 Opus$^\star$ & $\times$ & 23.5 \\
GPT-4-turbo-2024-04-09$^\star$ & $\times$ & 24.2 \\
Claude 3.5 Sonnet$^\star$ & $\times$ & 28.9 \\
GPT-4o~\citep{grok3_beta} & $\times$ & 38.2 \\
GPT-4o-2024-11-20$^\star$ & $\times$ & 38.8 \\
GPT-4o-2024-05-13$^\star$ & $\times$ & 39.0 \\
GPT-4o-2024-08-06$^\star$ & $\times$ & 40.1 \\
o1$^\star$ & $\times$ & 42.6 \\
o1-preview$^\star$ & $\times$ & 42.4 \\
Grok 3 Beta~\citep{grok3_beta} & $\times$ & 43.6 \\
Gemini 2.0 Pro~\citep{grok3_beta} & $\times$ & 44.3 \\
Claude-3.7-Sonnet~\citep{simpleqa_leaderboard} & $\times$ & 50.0 \\
Gemini 2.5 Pro~\citep{introducing_gemini25_pro} & $\times$ & 52.9 \\
GPT-4.5~\citep{introducing_gpt45} & $\times$ & 62.5 \\
Perplexity Sonar~\citep{perplexity_sonar_pro} & $\checkmark$ & 77.3 \\
Perplexity Sonar Reasoning Pro~\citep{perplexity_sonar_pro} & $\checkmark$ & 85.8 \\
Exa~\citep{exa_2025} & $\checkmark$ & 90.0 \\
Linkup Web Search~\citep{linkup_2025} & $\checkmark$ & 90.1 \\
Perplexity Deep Research~\citep{perplexity_deep_research} & $\checkmark$ & 93.9 \\
Liner Pro Reasoning~\citep{liner_pro_reasoning} & $\checkmark$ & 95.3 \\
\midrule
\rowcolor{mygreen}\multicolumn{3}{c}{\textit{Open‑source}} \\ \addlinespace
Llama-3.1-70B~\citep{alzubi2025open} & $\times$ & 20.4 \\
DeepSeek-V3~\citep{grok3_beta} & $\times$ & 24.9 \\
DeepSeek-R1~\citep{guo2025deepseek} & $\times$ & 30.1 \\
Open Deep Search (DeepSeek-R1) ~\citep{alzubi2025open} & $\checkmark$ & 88.3 \\
GLM-4.6 & $\checkmark$ & 91.9 \\
Kimi-Researcher$^\spadesuit$~\citep{moonshotai2025kimiresearcher} & $\checkmark$ & 93.6 \\
ROMA (GLM-4.6) & $\checkmark$ & \textbf{93.9} \\

\bottomrule
\end{tabular}
\end{table}
\vspace{-5pt}

%% file: tables/eqbench.tex
\begin{table}[t]
\centering
\caption{EQ-Bench Long-form writing scores (0--100), evaluated using Claude Sonnet 4 as the judge. $^\star$ denotes results taken from the EQ-Bench Long-form writing leaderboard.}
\vspace{5pt}
\label{table:eqbench_longform_writing}
\begin{tabular}{lc}
\toprule
\textbf{Model} & \textbf{Score (\%)} \\
\midrule
\rowcolor{myyellow}\multicolumn{2}{c}{\textit{Closed‑source}} \\ \addlinespace
Horizon Alpha$^\star$ & 70.0 \\
Gemini 3 Pro Preview$^\star$ & 70.6 \\

GPT-5 Chat$^\star$ & 70.6 \\
Gemini 2.5 Pro Preview-06-05$^\star$ & 70.6 \\
Claude 3.5 Sonnet$^\star$ & 70.9 \\
GPT-5-2025-08-07$^\star$ & 71.4 \\
Claude 3.7 Sonnet-2025-02-19$^\star$ & 71.6 \\
Gemini 2.5 Pro Preview-03-25$^\star$ & 72.0 \\
Claude Sonnet 4$^\star$ & 73.5 \\
Claude Opus 4.1$^\star$ & 74.1 \\
GPT-5.2$^\star$ & 74.5 \\
Claude Haiku 4.5$^\star$ & 76.5 \\
Polaris Alpha$^\star$ & 76.9 \\
Claude Opus 4.5-2025-11-01$^\star$ & 79.3 \\ 
Claude Sonnet 4.5$^\star$ & \textbf{79.8} \\
\midrule
\rowcolor{mygreen}\multicolumn{2}{c}{\textit{Open‑source}} \\ \addlinespace
GLM-4.6$^\star$ & 70.0 \\
DeepSeek-R1$^\star$ & 70.7 \\
DeepSeek V3.2$^\star$ & 72.6 \\
Kimi K2 Thinking$^\star$ & 73.1 \\
DeepSeek-V3.1$^\star$ & 73.6 \\
Kimi K2.5$^\star$ & 74.9 \\
Qwen3-Max-2025-09-24$^\star$ & 75.1 \\
ROMA w/ GEPA+ (DeepSeek-V3.1) & \textbf{79.8} \\
\bottomrule
\end{tabular}
\end{table}
\vspace{-5pt}

%% file: tables/eqbench_cost.tex
\begin{table}[t]
\centering
\caption{Average cost, token usage, and latency for ROMA (DeepSeek V3.1) in Chain-of-Thought mode with depth $=1$ on long-form writing tasks.}
\vspace{5pt}
\label{table:roma_cost_breakdown}
\begin{tabular}{lcccc}
\toprule
\textbf{Component} & \textbf{Cost (\$)} & \textbf{Input tokens} & \textbf{Output tokens} & \textbf{Latency (s)} \\
\midrule
Atomizer   & 0.01 & 17{,}709.64 &   129.86 &  3.61 \\
Planner    & 0.01 & 19{,}683.24 &   720.88 &  9.76 \\
Executor   & 0.00 &  3{,}554.45 & 1{,}002.11 & 12.18 \\
Aggregator & 0.02 & 33{,}476.38 & 1{,}220.21 & 19.12 \\
\midrule
\textbf{Total} & \textbf{0.05} & \textbf{74{,}423.71} & \textbf{3{,}073.06} & \textbf{44.67} \\
\bottomrule
\end{tabular}
\end{table}
\vspace{-5pt}

%% file: tables/abgen.tex
\begin{table}[t]
\centering
\caption{ROMA achieves the best Importance, Soundness, and Overall scores among both closed-source and open-source baselines on AbGen. Generation quality is evaluated using an LLM-as-a-judge setup with GPT-4.1-mini as the judge model.}
\vspace{5pt}
\label{table:abgen_results}
\begin{adjustbox}{max width=\textwidth}
\begin{tabular}{lcccc}
\toprule
\textbf{Model} & \textbf{Importance} & \textbf{Faithfulness} & \textbf{Soundness} & \textbf{Overall} \\
\midrule
\rowcolor{myyellow}\multicolumn{5}{c}{\textit{Closed‑source}} \\ \addlinespace
o4-mini & 4.66 & 4.59 & 4.27 & 4.78 \\
GPT-4o & 4.76 & 4.58 & 4.11 & 4.76 \\
GPT-4.1 & 4.78 & 4.72 & 4.28 & 4.86 \\
Gemini 2.5 Flash Preview-05-20 & 4.65 & 4.64 & 4.15 & 4.66 \\
\midrule
\rowcolor{mygreen}\multicolumn{5}{c}{\textit{Open‑source}} \\ \addlinespace
DeepSeek-R1-0528-Qwen3-8B & 4.59 & 4.50 & 4.10 & 4.61 \\
Llama-3.3-70B-Instruct-FP8-dynamic & 4.51 & 4.14 & 4.03 & 4.34 \\
Llama-4-Maverick-17B-128E-Instruct-FP8 & 4.50 & 4.42 & 4.03 & 4.56 \\
Llama-4-Scout-17B-16E-Instruct & 4.69 & 4.10 & 4.01 & 4.38 \\
Meta-Llama-3.1-70B-Instruct-FP8 & 4.67 & 4.15 & 4.04 & 4.40 \\
Mistral-Small-3.1-24B-Instruct-2503 & 4.67 & 4.40 & 4.10 & 4.66 \\
Qwen2.5-32B-Instruct & 4.67 & 4.17 & 4.07 & 4.61 \\
Qwen3-235B-A22B-fp8-tput & 4.69 & 4.58 & 4.18 & 4.84 \\
Qwen3-32B-AWQ & 4.72 & 4.43 & 4.19 & 4.73 \\
Qwen3-8B & 4.68 & 4.42 & 4.09 & 4.75 \\
DeepSeek-V3.2-Exp & 4.74 & \textbf{4.75} & 4.17 & 4.79 \\
deepseek-chat & 4.68 & 4.54 & 4.10 & 4.77 \\
deepseek-reasoner & 4.73 & 4.69 & 4.32 & 4.90 \\
gemma-3-27b-it & 4.58 & 4.53 & 4.20 & 4.59 \\
phi-4 & 4.55 & 4.42 & 4.09 & 4.61 \\
ROMA (DeepSeek-V3.2-Exp) & \textbf{4.91} & 4.64 & \textbf{4.62} & \textbf{4.93} \\
\bottomrule
\end{tabular}
\end{adjustbox}
\end{table}
\vspace{-5pt}

%% file: sections/related_work.tex
\section{Related Work}
\label{sec:related_work}
\paragraph{Hierarchical task decomposition:} Several works explore hierarchical decomposition to improve long-form generation. Early work in story and script generation adopts staged pipelines that plan high-level structure before realizing text, improving coherence over long horizons~\citep{fan-etal-2018-hierarchical,yao2019plan,mirowski2023co}. More recently, \cite{xiong-etal-2025-beyond} propose Heterogeneous Recursive Planning (HRP), which decomposes writing into typed subtasks (e.g., retrieval, reasoning, composition) executed by a task-specific scheduler inspired by Hierarchical Task Networks~\citep{sacerdoti1975structure,georgievski2015htn}. HRP assumes that each subtask can be assigned a single cognitive type, enabling targeted execution strategies. ROMA is complementary in spirit but differs structurally. Instead of embedding decomposition and scheduling logic into a task-specific planner, ROMA defines a uniform, recursive control loop (with explicit Atomizer, Planner, Executor, and Aggregator roles) that applies across domains and task types. Concurrent work on Recursive Language Models (RLMs)~\citep{zhang2025recursive} explores implicit recursion by allowing models to invoke themselves without an explicit meta-agent. In contrast, ROMA externalizes recursion into a transparent control abstraction, exposing task boundaries, dependencies, and aggregation steps as first-class execution traces.

\paragraph{Long-context limitations and context management: } Recent work shows that increasing context length alone does not yield reliable long-context reasoning. Empirical studies document strong positional biases~\citep{liu-etal-2024-lost,vu-etal-2024-freshllms} and performance degradation as inputs grow longer or more complex, a phenomenon referred to as context rot~\citep{hong2025context}. RULER~\citep{hsieh2024ruler} emphasizes failures in context utilization and multi-step reasoning, including multi-hop tracing and aggregation over long inputs. In parallel, specialized benchmarks for retrieval-augmented generation show that models struggle to reconcile noisy or conflicting evidence without explicit mechanisms for evidence selection and synthesis~\citep{pham2025sealqa,wang2025retrievalaugmented}. From a systems perspective, recent guidance argues that context must be treated as a precious, finite resource with diminishing marginal returns, which must be carefully managed within the model's context window~\citep{anthropic2025effective}. ROMA aligns with this perspective by structuring execution so that local computation and higher-level aggregation are separated, enabling intermediate compression without requiring all information to be present simultaneously.

\vspace{-5pt}
\paragraph{Test-time scaling and agentic frameworks:} Test-time scaling methods show that allocating more compute through \emph{parallel scaling} (e.g., majority voting or Best-of-N~\citep{snell2025scaling,brown2025large}, which generate multiple solution attempts in parallel and select the best one) or \emph{sequential scaling} (e.g., budget forcing~\citep{muennighoff-etal-2025-s1}, which iteratively refines solutions based on previous attempts) can improve performance, though gains are often limited by inefficient sampling strategies or fixed context windows. Building on these ideas, hybrid inference strategies interleave sequential reasoning with parallel exploration, including tree-based methods such as Monte Carlo Tree Search~\citep{zhang2023planning,zhou2024language} and guided beam search~\citep{xie2023selfevaluation}, as well as approaches that use process-level rewards to guide exploration and pruning~\citep{wu2024scaling}. These approaches are primarily evaluated in constrained settings such as math and coding, and their assumptions do not readily extend to general question answering or long-context scenarios. In parallel, agentic frameworks such as ReAct~\citep{yao2023react}, Tree-of-Thoughts~\citep{yao2023tree}, and Reflexion~\citep{shinn2023reflexion} interleave reasoning, action, and feedback during generation, typically with control logic embedded in prompts or bespoke orchestration code. ROMA complements these approaches by factoring planning, execution, and aggregation into explicit, reusable components with a shared execution protocol, supporting parallel execution, modular roles, and structured execution traces, while GEPA+ enables adaptation of component behavior within this framework.

%% file: sections/conclusions.tex
\section{Conclusions}
We present ROMA, a recursive, domain-agnostic meta-agent framework that addresses key structural limitations of existing agentic systems, including ad hoc orchestration, opaque execution, and uncontrolled context growth. By standardizing agent construction around a fixed recursive control loop with explicit roles for decomposition, execution, and aggregation, ROMA enables parallel, heterogeneous multi-agent systems while producing transparent, hierarchical execution traces. We further introduced GEPA+, a prompt optimization method tailored to ROMA's modular architecture, which enables efficient task adaptation without fine-tuning by jointly optimizing component prompts under interface constraints. Empirical results across reasoning and long-form generation benchmarks demonstrate that ROMA delivers strong system-level performance and can match or exceed competitive baselines, including specialized research agents and closed-source models. Taken as a whole, our results suggest that recursive, modular agent architectures provide a principled and practical foundation for building interpretable, flexible, and scalable agentic systems.

%% file: sections/limitations.tex
\section{Limitations and Future Work}

While ROMA provides a structured framework for building scalable agentic systems, it does not guarantee optimal task decomposition or aggregation. Atomization and MECE planning can fail, which may result in redundant subtrees, missing subgoals, or inefficient execution. Aggregation quality is similarly dependent on model behavior and prompt design, and may omit critical evidence or over-compress intermediate results, particularly under noisy or conflicting retrieval. Though parallel execution can reduce wall-clock latency, it may increase total compute cost and introduces practical coordination challenges, including tool budgets, rate limits, and failure recovery across concurrent branches. Finally, while ROMA mitigates context explosion by restricting local context and offloading artifacts, it does not eliminate long-context failure modes, and stronger task-aware verification and compression mechanisms remain necessary for high-stakes or safety-critical applications.

Future work includes improving decomposition and execution decisions through learned or verifier-guided planning objectives that explicitly trade off reasoning depth, cost, and uncertainty. ROMA's structured execution traces also enable trace-driven error localization and automated repair, such as targeted re-planning or selective re-execution of failed subtrees. Another promising direction is adaptive model routing, where models are chosen dynamically at each node based on expected value, latency, or reliability. Finally, extending ROMA to multimodal and safety-critical tool use will require richer notions of provenance, permissions, and execution guarantees, which we view as an important step toward deploying recursive agent architectures in high-stakes settings.

%% file: sections/appendices.tex
\appendix\section*{Appendix}
\section{Judging prompt}
\label{appendix:judging_prompt}
\begin{lstlisting}[style=llmstyle]
Judge whether the following [response] to [question] is correct or not based on the precise and unambiguous [correct_answer] below.

    [question]: {problem}

    [response]: {row["agent_answer"]}

    Your judgement must be in the format and criteria specified below:

    extracted_final_answer: The final exact answer extracted from the [response]. Put the extracted answer as 'None' if there is no exact, final answer to extract from the response.

    [correct_answer]: {row["answer"]}

    reasoning: Explain why the extracted_final_answer is correct or incorrect based on [correct_answer], focusing only on if there are meaningful differences between [correct_answer] and the extracted_final_answer. Do not comment on any background to the problem, do not attempt to solve the problem, do not argue for any answer different than [correct_answer], focus only on whether the answers match.

    correct: Answer 'yes' if extracted_final_answer matches the [correct_answer] given above, or is within a small margin of error for numerical problems. Answer 'no' otherwise, i.e. if there if there is any inconsistency, ambiguity, non-equivalency, or if the extracted answer is incorrect.

    confidence: The extracted confidence score between 0|\%| and 100|\%| from [response]. Put 100 if there is no confidence score available.
\end{lstlisting}
\section{ROMA's complete experimental configurations} \label{appendix:config_test}
\begin{lstlisting}[style = llmstyle]
agents:
  atomizer:
    llm:
      model: fireworks_ai/accounts/fireworks/models/glm-4p6
      temperature: 0.4
      max_tokens: 8000
    signature_instructions: "prompt_optimization.prompts.seed_prompts.atomizer_seed:ATOMIZER_PROMPT"
    demos: "prompt_optimization.prompts.seed_prompts.atomizer_seed:ATOMIZER_DEMOS"
    adapter_type: chat

  planner:
    llm:
      model: fireworks_ai/accounts/fireworks/models/glm-4p6
      temperature: 0.4
      max_tokens: 32000
    signature_instructions: "prompt_optimization.prompts.seed_prompts.planner_seed:PLANNER_PROMPT"
    demos: "prompt_optimization.prompts.seed_prompts.planner_seed:PLANNER_DEMOS"
    adapter_type: chat
    agent_config:
      max_subtasks: 12

    toolkits:
      - class_name: WebSearchToolkit
        enabled: true
        toolkit_config:
          model: openai/gpt-5-mini
          max_results: 5
          search_context_size: high
          temperature: 1.0  # Required for GPT-5 reasoning models
          max_tokens: 128_000  # Required for GPT-5 reasoning models

  aggregator:
    llm:
      model: fireworks_ai/accounts/fireworks/models/glm-4p6
      temperature: 0.6
      max_tokens: 180_000
    adapter_type: chat
    signature_instructions: "prompt_optimization.prompts.seed_prompts.aggregator_seed:AGGREGATOR_PROMPT"
    demos: "prompt_optimization.prompts.seed_prompts.aggregator_seed:AGGREGATOR_DEMOS"

  executors:
    RETRIEVE:
      llm:
        model: fireworks_ai/accounts/fireworks/models/glm-4p6  # Fast & cheap for simple queries
        temperature: 0.7  # Deterministic data retrieval
        max_tokens: 128_000  # Sufficient for data responses
      prediction_strategy: react
      signature_instructions: "prompt_optimization.prompts.seed_prompts.executor_retrieve_seed:EXECUTOR_RETRIEVE_PROMPT"
      demos: "prompt_optimization.prompts.seed_prompts.executor_retrieve_seed:EXECUTOR_RETRIEVE_DEMOS"
      agent_config:
        max_executions: 6  # Fewer iterations for simple retrievals
      adapter_type: chat
      toolkits:
        - class_name: WebSearchToolkit
          enabled: true
          toolkit_config:
            model: openai/gpt-5-mini
            # No search_engine = OpenRouter native search
            search_context_size: high
            temperature: 1.0  # Required for GPT-5 reasoning models
            max_tokens: 128_000  # Required for GPT-5 reasoning models
    THINK:
      llm:
        model: fireworks_ai/accounts/fireworks/models/glm-4p6  # Best reasoning
        temperature: 0.2  # Balanced creativity + accuracy
        max_tokens: 16000  # Large context for complex reasoning
      prediction_strategy: react
      adapter_type: chat
      signature_instructions: "prompt_optimization.prompts.seed_prompts.executor_think_seed:EXECUTOR_THINK_PROMPT"
      demos: "prompt_optimization.prompts.seed_prompts.executor_think_seed:EXECUTOR_THINK_DEMOS"
      agent_config:
        max_executions: 12  # Multiple iterations for deep analysis
      toolkits:
        - class_name: WebSearchToolkit
          enabled: true
          toolkit_config:
            model: openai/gpt-5-mini
            search_context_size: high
            temperature: 1.0  # Required for GPT-5 reasoning models
            max_tokens: 128_000  # Required for GPT-5 reasoning models

        - class_name: FileToolkit  # Save insights
          enabled: true
          toolkit_config:
            enable_delete: false
            max_file_size: 10485760  # HOMOGENIZED: 10MB standard

    WRITE:
      llm:
        model: fireworks_ai/accounts/fireworks/models/glm-4p6  # Best writing quality
        temperature: 0.3  # More creative for engaging writing
        max_tokens: 16000  # Long-form content
      prediction_strategy: react
      adapter_type: chat
      signature_instructions: "prompt_optimization.prompts.seed_prompts.executor_write_seed:EXECUTOR_WRITE_PROMPT"
      demos: "prompt_optimization.prompts.seed_prompts.executor_write_seed:EXECUTOR_WRITE_DEMOS"
      agent_config:
        max_executions: 8  # Moderate iterations for refinement
      toolkits:
        - class_name: FileToolkit  # PRIMARY: Save reports
          enabled: true
          toolkit_config:
            enable_delete: false
            max_file_size: 10485760  # HOMOGENIZED: 10MB standard
\end{lstlisting}


\section{GEPA+-optimized Atomizer prompt for long-form writing on EQ-Bench}
\label{appendix:gepa-plus-prompt-atomizer}
\begin{lstlisting}[style=llmstyle]
Atomizer - Instruction Prompt

Role
Classify the goal as ATOMIC or NOT and set `node_type`. Do not solve the task.

Available Executors (for atomic tasks only)
- Think, Search, Write

Decision Rules
- Atomic (->) EXECUTE) iff ALL are true:
  1) Single deliverable - exactly one answer/artefact/transformation.
  2) Single executor suffices - exactly one of Think OR Search OR Write can produce the final output in one pass.
  3) No inter-step dependencies - no "first do X then Y", no staged approvals, no prerequisite data collection, including implicit multi-hop reasoning where intermediate results are required.
  4) No multi-output packaging - not requesting multiple distinct artefacts or formats.
  5) No external coordination - no bookings, purchases, deployments, tests, or file operations.

Creative Writing Specific Rules
- Multi-chapter stories, novels, or serialized narratives -> PLAN (requires story structure, character arcs, plot development, world-building)
- Stories requiring "several chapters" or "multiple chapters" -> PLAN (explicitly multi-part)
- Character development, world-building, or plot planning tasks -> PLAN (foundational elements needed before writing)
- Single scene, single chapter, or very short creative piece (< 1000 words) -> May be ATOMIC if self-contained
- Creative writing prompts mentioning genre conventions, narrative structure, or story arcs -> PLAN (requires planning for coherence)

Notes
- Needing web retrieval or citations does not always force planning; if one Search pass can do it, it's atomic.
- Creative writing inherently benefits from planning: narrative structure, character consistency, plot coherence, pacing, and world-building all require decomposition.

When to choose PLAN (-> PLAN)
- Any multi-step sequencing (outline->draft, generate->evaluate->select, research A & B -> compare).
- Multiple deliverables or formats.
- Parallel subtasks to be synthesized.
- Clarification required before executing the goal.
- External actions/verification: bookings, deployments, tests, file or system operations.
- Long procedural projects with dependencies.
- Implicit multi-hop dependencies - chained lookups or intermediate computations are needed to reach the answer.
- Creative writing tasks requiring:
  * Multiple chapters or scenes
  * Character development and arcs
  * World-building and setting establishment
  * Plot structure and narrative progression
  * Genre-specific conventions and pacing
  * Story coherence across multiple sections

Tie-breaker
- If a single executor can reasonably deliver the end result in one pass, choose EXECUTE; otherwise PLAN.
- For creative writing: When in doubt, choose PLAN. Narrative structure, character consistency, and story coherence benefit from decomposition.

Strict Output Contract
- Return ONLY this JSON object (no prose, no extra keys, no markdown):
{
  "is_atomic": true|false,
  "node_type": "EXECUTE"|"PLAN"
}

Compliance
- Do not design plans, pick executors, or add explanations.
- Do not solve or partially solve the task.
- Output exactly the two fields above, nothing else.
\end{lstlisting}
\section{GEPA+-optimized Planner prompt for long-form writing on EQ-Bench}
\label{appendix:gepa-plus-prompt-planner}
\begin{lstlisting}[style=llmstyle]
Planner - Instruction Prompt

Role
Plan a goal into minimal, parallelizable subtasks with a precise, acyclic dependency graph. Do not execute; only plan.

Available Tools
If web search tools are available to you, you can use them during planning to:
- Research current events, trends, or market data when planning tasks that require up-to-date information
- Verify task requirements or gather context before decomposing complex goals
- Find relevant documentation, best practices, or domain-specific knowledge to inform your planning
- Improve the quality and accuracy of RETRIEVE task definitions
- Research genre conventions, narrative structures, or creative writing techniques for story planning

Output Contract (strict)
- Return only: `subtasks` and `dependencies_graph`. No extra keys, no prose.
- `subtasks`: list[SubTask]. Each SubTask MUST include:
  - `goal`: imperative, concrete objective for the subtask.
  - `task_type`: one of "THINK", "RETRIEVE", "WRITE".
  - `dependencies`: list[str] of subtask IDs it depends on.
  - `context_input` (optional): brief note on what to consume from dependencies; omit when unnecessary.
- `dependencies_graph`: dict[str, list[str]] | null
  - Keys and values are subtask IDs as 0-based indices encoded as strings, e.g., "0", "1".
  - Must be acyclic and consistent with each SubTask's `dependencies`.
  - Use empty lists for independent subtasks; set to `{}` if no dependencies, or `null` if not needed.
- Do not add fields like `id` or `result`. The list index is the subtask ID.

Task Type Guidance (MECE)
- THINK: reasoning, derivations, comparisons, validations; no external retrieval. For creative writing: story structure planning, character development, plot outlines, world-building concepts, genre analysis.
- RETRIEVE: fetch/verify external info where freshness, citations, or lookup are essential (replaces "SEARCH"). For creative writing: research genre conventions, historical settings, cultural details, or technical accuracy.
- WRITE: produce prose/structured text when inputs are known. For creative writing: chapters, scenes, character descriptions, world-building details, narrative prose.

Creative Writing Decomposition Strategy
For multi-chapter stories or complex narratives, decompose into:
1. Foundation Phase (THINK tasks, can be parallel):
   - Story structure and narrative arc (beginning, middle, end; three-act structure; chapter breakdown)
   - Character development (main characters, motivations, arcs, relationships)
   - World-building and setting (time, place, rules, atmosphere, genre conventions)
   - Plot outline (key events, conflicts, resolutions, pacing)
   - Genre-specific elements (tropes, conventions, tone, style)

2. Development Phase (WRITE tasks, depend on foundation):
   - Individual chapters or scenes (each chapter as separate WRITE task)
   - Character introductions and development scenes
   - World-building exposition and setting descriptions
   - Plot progression scenes (conflict, rising action, climax, resolution)

3. Synthesis Phase (WRITE task, depends on all chapters):
   - Final story assembly with smooth transitions
   - Consistency checks and narrative flow
   - Character voice consistency across chapters
   - Pacing and tension management

Story Structure Principles
- Narrative Arc: Establish beginning (setup, inciting incident) -> Middle (rising action, complications) -> End (climax, resolution)
- Chapter Structure: Each chapter should advance plot, develop characters, or build world; maintain chapter-to-chapter coherence
- Character Consistency: Character traits, voice, and motivations must remain consistent across all chapters
- Pacing: Balance action, dialogue, description, and introspection; vary chapter lengths and intensity
- Genre Awareness: Adapt structure to genre (mystery: clues and reveals; romance: emotional beats; horror: tension and dread)
- World-Building: Establish rules early; maintain internal consistency; show don't tell when possible

Decomposition Principles
- Minimality: Decompose only as much as necessary to reach the goal.
- MECE: Subtasks should not overlap; together they fully cover the goal.
- Parallelization: Foundation tasks (character development, world-building, plot outline) can often run in parallel; chapter writing depends on foundation but chapters can be written in parallel once foundation is set.
- Granularity: For creative writing, typical structure: 3-5 foundation tasks (THINK) -> 3-8 chapter tasks (WRITE) -> 1 synthesis task (WRITE). Total: 7-14 subtasks for multi-chapter stories.
- Determinism: Each subtask should have a clear, verifiable completion condition.

Dependency Rules
- Use 0-based indices as strings for IDs ("0", "1", ...). The index in `subtasks` is the ID.
- Foundation tasks (story structure, characters, world, plot) typically have no dependencies and can run in parallel.
- Chapter writing tasks depend on relevant foundation tasks (e.g., Chapter 1 depends on story structure, character development, world-building).
- Later chapters may depend on earlier chapters for narrative continuity (e.g., Chapter 2 depends on Chapter 1).
- Final synthesis depends on all chapter tasks.
- Keep the graph acyclic; avoid chains longer than necessary.
- Ensure `dependencies_graph` matches each SubTask's `dependencies` exactly.

Context Flow
- Outputs from dependencies are available to dependents; do not recompute.
- When a dependent needs specific artefacts (character descriptions, plot points, world details, previous chapter content), state this succinctly in `context_input`.
- For chapters: Reference character descriptions, plot outline, world-building details, and previous chapter content.

Edge Cases
- If the goal is already atomic, return the minimal valid plan (often 1-3 subtasks) rather than inflating to 3-8.
- If key requirements are unspecified, add an early THINK step to enumerate assumptions or a RETRIEVE step to collect missing facts.
- For very short stories (< 1000 words), may only need: story structure (THINK) -> single chapter (WRITE).
- For multi-chapter stories, always include foundation planning before chapter writing.

Strict Output Shape
{
  "subtasks": [SubTask, ...],
  "dependencies_graph": {"<id>": ["<id>", ...], ...} | {}
}

Do not execute any steps, and do not include reasoning or commentary in the output.
\end{lstlisting}
\section{GEPA+-optimized Executor prompt for long-form writing on EQ-Bench}
\label{appendix:gepa-plus-prompt-executor}
\begin{lstlisting}[style=llmstyle]
Executor (WRITE) - Instruction Prompt

Role
Execute WRITE tasks: create creative narrative prose, documentation, reports, summaries, and clear written communication across various formats and audiences.

Task Characteristics (WRITE)
- Primary goal: Produce clear, well-structured written content that serves its purpose
- Context-aware: Adapt approach based on task type (creative writing vs. technical/business writing)
- Audience-aware: Adapt tone, style, and complexity to target readers
- Purpose-driven: Content serves specific goal (entertain, inform, document, persuade, instruct)
- Format-flexible: Creative stories, technical docs, business reports, user guides, summaries, articles

Creative Writing Mode (for narrative prose, stories, chapters, scenes)
When writing creative fiction, narrative prose, or storytelling content:

Execution Guidelines (Creative Writing)
1. Narrative Voice: Establish and maintain consistent point of view (first, second, or third person) and narrative voice throughout
2. Show Don't Tell: Use sensory details, actions, dialogue, and scenes to reveal character and plot rather than exposition
3. Character Consistency: Maintain character traits, speech patterns, motivations, and relationships consistent with provided context
4. Scene Construction: Build scenes with clear setting, action, dialogue, and purpose; each scene should advance plot or develop character
5. Sensory Details: Engage all five senses (sight, sound, smell, taste, touch) to create immersive, vivid prose
6. Dialogue: Write natural, character-specific dialogue that reveals personality and advances plot; use dialogue tags sparingly
7. Pacing: Vary sentence length and structure; balance action, description, dialogue, and introspection; control narrative rhythm
8. Emotional Resonance: Evoke appropriate emotions through character reactions, sensory details, and narrative tension
9. Genre Awareness: Adapt style, tone, and conventions to genre (mystery, romance, horror, sci-fi, fantasy, etc.)
10. World-Building Integration: Seamlessly weave world-building details into narrative without info-dumping

Creative Writing Quality Standards
- Narrative Coherence: Story flows logically; plot points connect; no contradictions
- Character Development: Characters are believable, well-developed, and consistent
- Prose Quality: Writing is clear, engaging, well-crafted with strong imagery and voice
- Creativity: Story is original, imaginative, and compelling
- Pacing: Story maintains appropriate rhythm and momentum; tension builds effectively
- World-Building: Setting is well-established and internally consistent (if applicable)
- Emotional Engagement: Writing evokes appropriate emotions and reader connection
- Faithfulness to Prompt: Story addresses the given writing prompt and genre requirements

Creative Writing Techniques
- Opening Hooks: Start with action, dialogue, or intriguing detail to engage reader immediately
- Scene Transitions: Use smooth transitions between scenes and chapters; maintain narrative flow
- Character Voice: Each character should have distinct voice, speech patterns, and mannerisms
- Subtext: Layer meaning beneath surface dialogue and action; let readers infer
- Foreshadowing: Plant subtle clues and hints for later plot developments
- Conflict: Every scene should contain conflict, tension, or stakes (internal or external)
- Description: Use specific, concrete details rather than vague abstractions
- Metaphor and Imagery: Employ figurative language to create vivid mental pictures
- Pacing Variation: Alternate fast-paced action with slower, reflective moments
- Chapter Endings: End chapters with hooks, cliffhangers, or emotional beats to maintain reader interest

Technical/Business Writing Mode (for documentation, reports, summaries)
When writing non-fiction, technical, or business content:

Execution Guidelines (Technical/Business Writing)
1. Audience analysis: Understand who will read this and what they need
2. Purpose clarity: Define what the document should accomplish
3. Structure first: Outline before writing (hierarchy, sections, flow)
4. Clarity over cleverness: Simple, direct language; avoid unnecessary jargon
5. Active voice: "The system processes requests" not "Requests are processed by the system"
6. Concrete examples: Use specific examples, not abstract descriptions
7. Visual hierarchy: Use headings, bullets, tables, and formatting for scanability

Technical/Business Quality Standards
- Clarity: Complex ideas explained simply; no ambiguity
- Completeness: All necessary information included; no critical gaps
- Accuracy: Facts verified, claims supported, technical details correct
- Organization: Logical flow, clear sections, easy navigation
- Conciseness: No unnecessary words; respect reader's time
- Consistency: Terminology, style, and formatting uniform throughout

Output Contract (strict)
- `output` (string): Well-structured written content in appropriate format (creative prose or technical/business content)
- `sources` (list[str]): Reference materials, data sources, documentation, or creative inspirations consulted

Context Integration (Critical for Creative Writing)
- Character Consistency: If writing a chapter in a multi-chapter story, maintain character descriptions, traits, and voice from previous chapters
- Plot Continuity: Build on previous plot developments; maintain narrative coherence across chapters
- World-Building Consistency: Adhere to established world rules, setting details, and genre conventions from foundation planning
- Tone and Style: Match the genre tone, narrative voice, and writing style established in the story structure
- Chapter Transitions: If writing a later chapter, reference events and character states from earlier chapters naturally

Common WRITE Patterns
Creative Writing:
- Chapter Writing: Establish scene -> Introduce/develop characters -> Advance plot -> Build tension -> End with hook
- Scene Writing: Set scene -> Character action/dialogue -> Conflict/tension -> Resolution or escalation -> Transition
- Character Introduction: Show character through action -> Reveal through dialogue -> Establish motivation -> Create connection

Technical/Business Writing:
- Technical documentation: Define scope -> Explain concepts -> Provide examples -> Document edge cases -> Include troubleshooting
- Business report: Executive summary -> Background/context -> Analysis -> Findings -> Recommendations
- User guide: Overview -> Prerequisites -> Step-by-step instructions -> Screenshots/examples -> FAQs
- API documentation: Endpoint description -> Parameters -> Request/response examples -> Error codes
- Summary: Read source -> Extract key points -> Organize logically -> Write concisely -> Verify accuracy

Writing Best Practices
Creative Writing:
- Start in media res (in the middle of action) when possible
- Use specific, concrete nouns and verbs; avoid weak modifiers
- Vary sentence structure for rhythm and emphasis
- Let characters reveal themselves through actions and dialogue
- Build tension through conflict, obstacles, and stakes
- Use setting to reflect mood and character state
- End scenes and chapters with forward momentum

Technical/Business Writing:
- Lead with most important information (inverted pyramid)
- One idea per paragraph
- Use transition words for flow (however, therefore, additionally)
- Define acronyms on first use
- Include code examples for technical content
- Add tables for comparison or reference data
- Use numbered lists for sequences, bullets for unordered items
- End with clear next steps or conclusion

Format-Specific Guidelines
Creative Writing:
- Chapter: Clear chapter heading, scene breaks, consistent POV, narrative flow, chapter-ending hook
- Scene: Setting establishment, character action, dialogue, conflict, resolution or escalation
- Short Story: Complete narrative arc, character development, satisfying conclusion

Technical/Business Writing:
- README: Purpose, installation, quickstart, examples, contributing
- API docs: Authentication, endpoints, parameters, responses, errors
- Changelog: Version, date, added/changed/fixed/deprecated/removed
- User guide: Task-oriented, step-by-step, screenshots, troubleshooting
- Report: Executive summary, methodology, findings, recommendations
- Tutorial: Learning objectives, prerequisites, incremental steps, exercises
\end{lstlisting}
\section{GEPA+ configuration details}
\label{appendix:gepa-plus-config}
\begin{lstlisting}[style=llmstyle]
{
  "gepa_plus": {
    "proposal_lms": [
      {
        "model": "openai/gpt-5",
        "temperature": 1.0,
        "max_tokens": 16000,
        "role": "proposal"
      },
      {
        "model": "openrouter/anthropic/claude-sonnet-4.5",
        "temperature": 1.0,
        "max_tokens": 16000,
        "role": "proposal"
      },
      {
        "model": "openrouter/google/gemini-2.5-flash",
        "temperature": 0.6,
        "max_tokens": 16000,
        "role": "proposal"
      }
    ],
    "judge_lm": {
      "model": "openrouter/anthropic/claude-sonnet-4.5",
      "temperature": 1.0,
      "max_tokens": 16000
    },
    "merger_lm": {
      "model": "openai/gpt-5",
      "temperature": 1.0,
      "max_tokens": 16000
    },
    "top_n": 2,
    "verbose": true
  }
}
\end{lstlisting}













\newpage
\section{Comparison of GEPA and GEPA+}
\label{appendix:gepa_vs_gepa_plus}
\input{tables/gepa_plus_performance}
\input{tables/gepa_plus_efficienty}

We evaluate GEPA+ against standard GEPA on three representative benchmarks that span distinct evaluation regimes: AIME25 (mathematical reasoning),\footnote{\url{https://artofproblemsolving.com/wiki/index.php/2025_AIME_I}} HotpotQA~\citep{yang-etal-2018-hotpotqa} (multi-hop question answering), and PAPILLON~\citep{gong2025papillon} (safety-oriented evaluation). Across all experiments, we use GPT-4.1-mini as the base model for evaluation, with PAPILLON additionally involving GPT-4.1-nano in its delegation setting, following the benchmark's design. Across all benchmarks, GEPA+ consistently improves both final accuracy and optimization efficiency, demonstrating more effective prompt adaptation under a fixed compute budget.

\paragraph{Performance: } Table~\ref{table:gepa_plus_performance} summarizes average accuracy over five runs. GEPA+ consistently outperforms GEPA across all three datasets. On AIME25, GEPA+ improves accuracy from 30.0\% to 32.3\%. On HotpotQA, performance increases from 62.0\% to 65.0\%. On PAPILLON, GEPA+ achieves 91.7\% accuracy, compared to 85.6\% for GEPA. These gains demonstrate that GEPA+ delivers reliable improvements across both reasoning-focused and safety-related tasks.

\paragraph{Efficiency: } Beyond accuracy, GEPA+ substantially improves optimization efficiency. We analyze convergence behavior on AIME25, where optimization cost is dominated by expensive metric evaluations (since each accuracy estimate requires re-running the model over many reasoning-intensive problems with full inference). As shown in Table~\ref{table:gepa_plus_efficiency}, standard GEPA requires 560 metric calls to converge, whereas GEPA+ converges using only 150 metric calls, a reduction of approximately 73\%. When normalized by metric calls, GEPA+ achieves 1.75$\times$ higher efficiency in terms of accuracy improvement per call. This reduction directly translates into lower compute cost and faster iteration.

Overall, these results show that GEPA+ not only improves downstream accuracy, but also achieves faster and more compute-efficient convergence. This makes GEPA+ particularly well-suited for optimizing multi-component agent systems, where evaluation costs are high and repeated optimization cycles are required.

%% file: tables/gepa_plus_performance.tex
\begin{table}[t]
\centering
\caption{Performance of GEPA and GEPA+ across datasets (averaged over five runs). All datasets use GPT-4.1-mini as the base evaluation model, with PAPILLON additionally involving GPT-4.1-nano in a privacy-preserving delegation setting where a small local model (GPT-4.1-nano) leverages a larger untrusted model (GPT-4.1-mini), following the benchmark's design.}
\vspace{5pt}
\label{table:gepa_plus_performance}
\begin{tabular}{lccccc}
\toprule
\textbf{Prompt optimizer} & \textbf{AIME25 (\%)} & \textbf{HotpotQA (\%)} & \textbf{PAPILLON (\%)} \\
\midrule
w/ GEPA  & 30.0 & 62.0 & 85.6  \\
w/ GEPA+ & 32.3 & 65.0 & 91.7  \\
\bottomrule
\end{tabular}
\end{table}
\vspace{-5pt}

%% file: tables/gepa_plus_efficienty.tex
\begin{table}[t]
\centering
\caption{Optimization efficiency comparison on AIME.}
\vspace{5pt}
\label{table:gepa_plus_efficiency}
\begin{tabular}{lccc}
\toprule
\textbf{Method} & \textbf{Metric calls} & \textbf{Efficiency (\% / call)} & \textbf{Relative efficiency} \\
\midrule
GEPA  & 560 & 0.0179 & 1.0$\times$ \\
GEPA+ & 150 & 0.0313 & 1.75$\times$ \\
\bottomrule
\end{tabular}
\end{table}
\vspace{-5pt}

%% file: neurips_2025.bib
@misc{anthropic2025claudecode,
    author       = {Anthropic},
    title        = {Claude Code: Best practices for agentic coding},
    year         = {2025},
    howpublished = {\url{https://www.anthropic.com/engineering/claude-code-best-practices/}},
}

@misc{cursor2026,
    author       = {Cursor},
    title        = {Best practices for coding with agents},
    year         = {2026},
    howpublished = {\url{https://cursor.com/blog/agent-best-practices/}},
}

@misc{openai_deep_research,
    author       = {Open{AI}},
    title        = {Introducing deep research},
    year         = {2025},
    howpublished = {\url{https://openai.com/index/introducing-deep-research/}},
}

@misc{perplexity_deep_research,
    author       = {Perplexity{AI}},
    title        = {Introducing Perplexity Deep Research},
    year         = {2025},
    howpublished = {\url{https://www.perplexity.ai/hub/blog/introducing-perplexity-deep-research/}},
}

@article{guo2024large,
    title={Large language model based multi-agents: A survey of progress and challenges},
    author={Guo, Taicheng and Chen, Xiuying and Wang, Yaqi and Chang, Ruidi and Pei, Shichao and Chawla, Nitesh V and Wiest, Olaf and Zhang, Xiangliang},
    journal={arXiv preprint arXiv:2402.01680},
    url={https://arxiv.org/abs/2402.01680},
    year={2024}
}

@inproceedings{xiong-etal-2025-beyond,
    title = "Beyond Outlining: Heterogeneous Recursive Planning for Adaptive Long-form Writing with Language Models",
    author = {Xiong, Ruibin  and
      Chen, Yimeng  and
      Khizbullin, Dmitrii  and
      Zhuge, Mingchen  and
      Schmidhuber, J{\"u}rgen},
    editor = "Christodoulopoulos, Christos  and
      Chakraborty, Tanmoy  and
      Rose, Carolyn  and
      Peng, Violet",
    booktitle = "Proceedings of the 2025 Conference on Empirical Methods in Natural Language Processing",
    month = nov,
    year = "2025",
    address = "Suzhou, China",
    publisher = "Association for Computational Linguistics",
    url = "https://aclanthology.org/2025.emnlp-main.1254/",
    doi = "10.18653/v1/2025.emnlp-main.1254",
    pages = "24678--24714",
    ISBN = "979-8-89176-332-6",
}

@article{agrawal2025gepa,
    title={Gepa: Reflective prompt evolution can outperform reinforcement learning},
    author={Agrawal, Lakshya A and Tan, Shangyin and Soylu, Dilara and Ziems, Noah and Khare, Rishi and Opsahl-Ong, Krista and Singhvi, Arnav and Shandilya, Herumb and Ryan, Michael J and Jiang, Meng and others},
    journal={arXiv preprint arXiv:2507.19457},
    url={https://arxiv.org/abs/2507.19457},
    year={2025}
}

@techreport{hong2025context,
    title = {Context Rot: How Increasing Input Tokens Impacts LLM Performance},
    author = {Hong, Kelly and Troynikov, Anton and Huber, Jeff},
    year = {2025},
    month = {July},
    institution = {Chroma},
    url = {https://research.trychroma.com/context-rot},
}

@article{pham2025sealqa,
    title={SealQA: Raising the Bar for Reasoning in Search-Augmented Language Models},
    author={Pham, Thinh and Nguyen, Nguyen and Zunjare, Pratibha and Chen, Weiyuan and Tseng, Yu-Min and Vu, Tu},
    journal={arXiv preprint arXiv:2506.01062},
    url={https://arxiv.org/abs/2506.01062},
    year={2025}
}

@inproceedings{krishna-etal-2025-fact,
    title = "Fact, Fetch, and Reason: A Unified Evaluation of Retrieval-Augmented Generation",
    author = "Krishna, Satyapriya  and
      Krishna, Kalpesh  and
      Mohananey, Anhad  and
      Schwarcz, Steven  and
      Stambler, Adam  and
      Upadhyay, Shyam  and
      Faruqui, Manaal",
    editor = "Chiruzzo, Luis  and
      Ritter, Alan  and
      Wang, Lu",
    booktitle = "Proceedings of the 2025 Conference of the Nations of the Americas Chapter of the Association for Computational Linguistics: Human Language Technologies (Volume 1: Long Papers)",
    month = apr,
    year = "2025",
    address = "Albuquerque, New Mexico",
    publisher = "Association for Computational Linguistics",
    url = "https://aclanthology.org/2025.naacl-long.243/",
    doi = "10.18653/v1/2025.naacl-long.243",
    pages = "4745--4759",
    ISBN = "979-8-89176-189-6",
}

@article{wei2024measuring,
    title={Measuring short-form factuality in large language models},
    author={Wei, Jason and Karina, Nguyen and Chung, Hyung Won and Jiao, Yunxin Joy and Papay, Spencer and Glaese, Amelia and Schulman, John and Fedus, William},
    journal={arXiv preprint arXiv:2411.04368},
    url={https://arxiv.org/abs/2411.04368},
    year={2024}
}

@article{paech2023eq,
    title={Eq-bench: An emotional intelligence benchmark for large language models},
    author={Paech, Samuel J},
    journal={arXiv preprint arXiv:2312.06281},
    url={https://arxiv.org/abs/2312.06281},
    year={2023}
}

@inproceedings{zhao-etal-2025-abgen,
    title = "{A}b{G}en: Evaluating Large Language Models in Ablation Study Design and Evaluation for Scientific Research",
    author = "Zhao, Yilun  and
      Chen, Weiyuan  and
      Xu, Zhijian  and
      Patwardhan, Manasi  and
      Wang, Chengye  and
      Liu, Yixin  and
      Vig, Lovekesh  and
      Cohan, Arman",
    editor = "Che, Wanxiang  and
      Nabende, Joyce  and
      Shutova, Ekaterina  and
      Pilehvar, Mohammad Taher",
    booktitle = "Proceedings of the 63rd Annual Meeting of the Association for Computational Linguistics (Volume 1: Long Papers)",
    month = jul,
    year = "2025",
    address = "Vienna, Austria",
    publisher = "Association for Computational Linguistics",
    url = "https://aclanthology.org/2025.acl-long.611/",
    doi = "10.18653/v1/2025.acl-long.611",
    pages = "12479--12491",
    ISBN = "979-8-89176-251-0",
}

@article{zeng2025glm,
    title={Glm-4.5: Agentic, reasoning, and coding (arc) foundation models},
    author={Zeng, Aohan and Lv, Xin and Zheng, Qinkai and Hou, Zhenyu and Chen, Bin and Xie, Chengxing and Wang, Cunxiang and Yin, Da and Zeng, Hao and Zhang, Jiajie and others},
    journal={arXiv preprint arXiv:2508.06471},
    url={https://arxiv.org/abs/2508.06471},
    year={2025}
}

@misc{moonshotai2025kimiresearcher,
    title={{K}imi-{R}esearcher: {E}nd-to-End {RL} Training for Emerging Agentic Capabilities}, 
    author={Moonshot AI},
    year         = {2025},
    howpublished = {\url{https://moonshotai.github.io/Kimi-Researcher/}},
}

@article{liu2024deepseek,
    title={Deepseek-v3 technical report},
    author={Liu, Aixin and Feng, Bei and Xue, Bing and Wang, Bingxuan and Wu, Bochao and Lu, Chengda and Zhao, Chenggang and Deng, Chengqi and Zhang, Chenyu and Ruan, Chong and others},
    journal={arXiv preprint arXiv:2412.19437},
    url={https://arxiv.org/abs/2412.19437},
    year={2024}
}

@misc{anthropic2025introducing,
    title={{I}ntroducing {C}laude {S}onnet 4.5}, 
    author={Anthropic},
    year         = {2025},
    howpublished = {\url{https://www.anthropic.com/news/claude-sonnet-4-5}},
}

@inproceedings{
    yao2023react,
    title={ReAct: Synergizing Reasoning and Acting in Language Models},
    author={Shunyu Yao and Jeffrey Zhao and Dian Yu and Nan Du and Izhak Shafran and Karthik R Narasimhan and Yuan Cao},
    booktitle={The Eleventh International Conference on Learning Representations },
    year={2023},
    url={https://openreview.net/forum?id=WE_vluYUL-X}
}

@inproceedings{
    wang2024executable,
    title={Executable Code Actions Elicit Better {LLM} Agents},
    author={Xingyao Wang and Yangyi Chen and Lifan Yuan and Yizhe Zhang and Yunzhu Li and Hao Peng and Heng Ji},
    booktitle={Forty-first International Conference on Machine Learning},
    year={2024},
    url={https://openreview.net/forum?id=jJ9BoXAfFa}
}

@inproceedings{
    wei2022chain,
    title={Chain of Thought Prompting Elicits Reasoning in Large Language Models},
    author={Jason Wei and Xuezhi Wang and Dale Schuurmans and Maarten Bosma and brian ichter and Fei Xia and Ed H. Chi and Quoc V Le and Denny Zhou},
    booktitle={Advances in Neural Information Processing Systems},
    editor={Alice H. Oh and Alekh Agarwal and Danielle Belgrave and Kyunghyun Cho},
    year={2022},
    url={https://openreview.net/forum?id=_VjQlMeSB_J}
}

@inproceedings{
    khattab2024dspy,
    title={{DSP}y: Compiling Declarative Language Model Calls into State-of-the-Art Pipelines},
    author={Omar Khattab and Arnav Singhvi and Paridhi Maheshwari and Zhiyuan Zhang and Keshav Santhanam and Sri Vardhamanan A and Saiful Haq and Ashutosh Sharma and Thomas T. Joshi and Hanna Moazam and Heather Miller and Matei Zaharia and Christopher Potts},
    booktitle={The Twelfth International Conference on Learning Representations},
    year={2024},
    url={https://openreview.net/forum?id=sY5N0zY5Od}
}

@misc{anthropic2024introducing,
    author       = {Anthropic},
    title        = {Introducing the Model Context Protocol},
    year         = {2024},
    howpublished = {\url{https://www.anthropic.com/news/model-context-protocol/}},
}

@inproceedings{
    zheng2023judging,
    title={Judging {LLM}-as-a-Judge with {MT}-Bench and Chatbot Arena},
    author={Lianmin Zheng and Wei-Lin Chiang and Ying Sheng and Siyuan Zhuang and Zhanghao Wu and Yonghao Zhuang and Zi Lin and Zhuohan Li and Dacheng Li and Eric Xing and Hao Zhang and Joseph E. Gonzalez and Ion Stoica},
    booktitle={Thirty-seventh Conference on Neural Information Processing Systems Datasets and Benchmarks Track},
    year={2023},
    url={https://openreview.net/forum?id=uccHPGDlao}
}

@article{liu-etal-2024-lost,
    title = "Lost in the Middle: How Language Models Use Long Contexts",
    author = "Liu, Nelson F.  and
      Lin, Kevin  and
      Hewitt, John  and
      Paranjape, Ashwin  and
      Bevilacqua, Michele  and
      Petroni, Fabio  and
      Liang, Percy",
    journal = "Transactions of the Association for Computational Linguistics",
    volume = "12",
    year = "2024",
    address = "Cambridge, MA",
    publisher = "MIT Press",
    url = "https://aclanthology.org/2024.tacl-1.9/",
    doi = "10.1162/tacl_a_00638",
    pages = "157--173",
}

@inproceedings{vu-etal-2024-freshllms,
    title = "{F}resh{LLM}s: Refreshing Large Language Models with Search Engine Augmentation",
    author = "Vu, Tu  and
      Iyyer, Mohit  and
      Wang, Xuezhi  and
      Constant, Noah  and
      Wei, Jerry  and
      Wei, Jason  and
      Tar, Chris  and
      Sung, Yun-Hsuan  and
      Zhou, Denny  and
      Le, Quoc  and
      Luong, Thang",
    editor = "Ku, Lun-Wei  and
      Martins, Andre  and
      Srikumar, Vivek",
    booktitle = "Findings of the Association for Computational Linguistics: ACL 2024",
    month = aug,
    year = "2024",
    address = "Bangkok, Thailand",
    publisher = "Association for Computational Linguistics",
    url = "https://aclanthology.org/2024.findings-acl.813/",
    doi = "10.18653/v1/2024.findings-acl.813",
    pages = "13697--13720",
}

@inproceedings{
    hsieh2024ruler,
    title={{RULER}: What{\textquoteright}s the Real Context Size of Your Long-Context Language Models?},
    author={Cheng-Ping Hsieh and Simeng Sun and Samuel Kriman and Shantanu Acharya and Dima Rekesh and Fei Jia and Boris Ginsburg},
    booktitle={First Conference on Language Modeling},
    year={2024},
    url={https://openreview.net/forum?id=kIoBbc76Sy}
    }

@inproceedings{
    wang2025retrievalaugmented,
    title={Retrieval-Augmented Generation with Conflicting Evidence},
    author={Han Wang and Archiki Prasad and Elias Stengel-Eskin and Mohit Bansal},
    booktitle={Second Conference on Language Modeling},
    year={2025},
    url={https://openreview.net/forum?id=z1MHB2m3V9}
}

@misc{anthropic2025effective,
    title={Effective context engineering for AI agents}, 
    author={Anthropic},
    year         = {2025},
    howpublished = {\url{https://www.anthropic.com/engineering/effective-context-engineering-for-ai-agents}},
}

@inproceedings{muennighoff-etal-2025-s1,
    title = "s1: Simple test-time scaling",
    author = "Muennighoff, Niklas  and
      Yang, Zitong  and
      Shi, Weijia  and
      Li, Xiang Lisa  and
      Fei-Fei, Li  and
      Hajishirzi, Hannaneh  and
      Zettlemoyer, Luke  and
      Liang, Percy  and
      Candes, Emmanuel  and
      Hashimoto, Tatsunori",
    editor = "Christodoulopoulos, Christos  and
      Chakraborty, Tanmoy  and
      Rose, Carolyn  and
      Peng, Violet",
    booktitle = "Proceedings of the 2025 Conference on Empirical Methods in Natural Language Processing",
    month = nov,
    year = "2025",
    address = "Suzhou, China",
    publisher = "Association for Computational Linguistics",
    url = "https://aclanthology.org/2025.emnlp-main.1025/",
    doi = "10.18653/v1/2025.emnlp-main.1025",
    pages = "20275--20321",
    ISBN = "979-8-89176-332-6"
}

@inproceedings{
    snell2025scaling,
    title={Scaling {LLM} Test-Time Compute Optimally Can be More Effective than Scaling Parameters for Reasoning},
    author={Charlie Victor Snell and Jaehoon Lee and Kelvin Xu and Aviral Kumar},
    booktitle={The Thirteenth International Conference on Learning Representations},
    year={2025},
    url={https://openreview.net/forum?id=4FWAwZtd2n}
}

@misc{
    brown2025large,
    title={Large Language Monkeys: Scaling Inference Compute with Repeated Sampling},
    author={Bradley Brown and Jordan Juravsky and Ryan Saul Ehrlich and Ronald Clark and Quoc V Le and Christopher Re and Azalia Mirhoseini},
    year={2025},
    url={https://openreview.net/forum?id=0xUEBQV54B}
}

@inproceedings{
    zhang2023planning,
    title={Planning with Large Language Models for Code Generation},
    author={Shun Zhang and Zhenfang Chen and Yikang Shen and Mingyu Ding and Joshua B. Tenenbaum and Chuang Gan},
    booktitle={The Eleventh International Conference on Learning Representations },
    year={2023},
    url={https://openreview.net/forum?id=Lr8cOOtYbfL}
}

@inproceedings{
    xie2023selfevaluation,
    title={Self-Evaluation Guided Beam Search for Reasoning},
    author={Yuxi Xie and Kenji Kawaguchi and Yiran Zhao and Xu Zhao and Min-Yen Kan and Junxian He and Qizhe Xie},
    booktitle={Thirty-seventh Conference on Neural Information Processing Systems},
    year={2023},
    url={https://openreview.net/forum?id=Bw82hwg5Q3}
}

@inproceedings{
    yao2023tree,
    title={Tree of Thoughts: Deliberate Problem Solving with Large Language Models},
    author={Shunyu Yao and Dian Yu and Jeffrey Zhao and Izhak Shafran and Thomas L. Griffiths and Yuan Cao and Karthik R Narasimhan},
    booktitle={Thirty-seventh Conference on Neural Information Processing Systems},
    year={2023},
    url={https://openreview.net/forum?id=5Xc1ecxO1h}
}

@inproceedings{
    shinn2023reflexion,
    title={Reflexion: language agents with verbal reinforcement learning},
    author={Noah Shinn and Federico Cassano and Ashwin Gopinath and Karthik R Narasimhan and Shunyu Yao},
    booktitle={Thirty-seventh Conference on Neural Information Processing Systems},
    year={2023},
    url={https://openreview.net/forum?id=vAElhFcKW6}
}

@inproceedings{
    zhou2024language,
    title={Language Agent Tree Search Unifies Reasoning, Acting, and Planning in Language Models},
    author={Andy Zhou and Kai Yan and Michal Shlapentokh-Rothman and Haohan Wang and Yu-Xiong Wang},
    booktitle={Forty-first International Conference on Machine Learning},
    year={2024},
    url={https://openreview.net/forum?id=njwv9BsGHF}
}

@inproceedings{
    wu2024scaling,
    title={Scaling Inference Computation: Compute-Optimal Inference for Problem-Solving with Language Models},
    author={Yangzhen Wu and Zhiqing Sun and Shanda Li and Sean Welleck and Yiming Yang},
    booktitle={The 4th Workshop on Mathematical Reasoning and AI at NeurIPS'24},
    year={2024},
    url={https://openreview.net/forum?id=j7DZWSc8qu}
}

@inproceedings{yao2019plan,
    title={Plan-and-write: Towards better automatic storytelling},
    author={Yao, Lili and Peng, Nanyun and Weischedel, Ralph and Knight, Kevin and Zhao, Dongyan and Yan, Rui},
    booktitle={Proceedings of the AAAI Conference on Artificial Intelligence},
    volume={33},
    pages={7378--7385},
    year={2019}
}

@inproceedings{fan-etal-2018-hierarchical,
    title = "Hierarchical Neural Story Generation",
    author = "Fan, Angela  and
      Lewis, Mike  and
      Dauphin, Yann",
    editor = "Gurevych, Iryna  and
      Miyao, Yusuke",
    booktitle = "Proceedings of the 56th Annual Meeting of the Association for Computational Linguistics (Volume 1: Long Papers)",
    month = jul,
    year = "2018",
    address = "Melbourne, Australia",
    publisher = "Association for Computational Linguistics",
    url = "https://aclanthology.org/P18-1082/",
    doi = "10.18653/v1/P18-1082",
    pages = "889--898",
}

@inproceedings{mirowski2023co,
    title={Co-writing screenplays and theatre scripts with language models: Evaluation by industry professionals},
    author={Mirowski, Piotr and Mathewson, Kory W and Pittman, Jaylen and Evans, Richard},
    booktitle={Proceedings of the 2023 CHI conference on human factors in computing systems},
    pages={1--34},
    year={2023}
}

@article{vu2024foundational,
  title        = {Foundational Autoraters: Taming Large Language Models for Better Automatic Evaluation},
  author       = {Vu, Tu and Krishna, Kalpesh and Alzubi, Salaheddin and Tar, Chris and Faruqui, Manaal and Sung, Yun-Hsuan},
  year         = {2024},
  journal      = {arXiv preprint arXiv:2407.10817},
  url          = {https://arxiv.org/abs/2407.10817}
}

@article{alzubi2025opendeepsearch2,
  title        = {Open Deep Search: Democratizing Search with Open-Source Reasoning Agents},
  author       = {Alzubi, Salaheddin and Brooks, Creston and Chiniya, Purva and Contente, Edoardo and von Gerlach, Chiara and Irwin, Lucas and Jiang, Yihan and Kaz, Arda and Nguyen, Windsor and Oh, Sewoong and Tyagi, Himanshu and Viswanath, Pramod},
  year         = {2025},
  journal      = {arXiv preprint arXiv:2503.20201},
  url          = {https://arxiv.org/abs/2503.20201}
}

@article{sacerdoti1975structure,
    title={A structure for plans and behavior},
    journal = {DTIC},
    author={Sacerdoti, Earl D},
    year={1975}
}

@article{georgievski2015htn,
    title={HTN planning: Overview, comparison, and beyond},
    author={Georgievski, Ilche and Aiello, Marco},
    journal={Artificial Intelligence},
    volume={222},
    pages={124--156},
    year={2015},
    publisher={Elsevier}
}

@article{zhang2025recursive,
    title={Recursive Language Models},
    author={Zhang, Alex L and Kraska, Tim and Khattab, Omar},
    journal={arXiv preprint arXiv:2512.24601},
    url={https://arxiv.org/abs/2512.24601},
    year={2025}
}

@article{singh2025openai,
    title={Openai gpt-5 system card},
    author={Singh, Aaditya and Fry, Adam and Perelman, Adam and Tart, Adam and Ganesh, Adi and El-Kishky, Ahmed and McLaughlin, Aidan and Low, Aiden and Ostrow, AJ and Ananthram, Akhila and others},
    journal={arXiv preprint arXiv:2601.03267},
    url={https://arxiv.org/abs/2601.03267},
    year={2025}
}

@article{comanici2025gemini,
    title={Gemini 2.5: Pushing the frontier with advanced reasoning, multimodality, long context, and next generation agentic capabilities},
    author={Comanici, Gheorghe and Bieber, Eric and Schaekermann, Mike and Pasupat, Ice and Sachdeva, Noveen and Dhillon, Inderjit and Blistein, Marcel and Ram, Ori and Zhang, Dan and Rosen, Evan and others},
    journal={arXiv preprint arXiv:2507.06261},
    url={https://arxiv.org/abs/2507.06261},
    year={2025}
}

@misc{perplexity_sonar_pro,
    author       = {Perplexity{AI}},
    title        = {Introducing the Sonar Pro API },
    year         = {2025},
    howpublished = {\url{https://www.perplexity.ai/hub/blog/introducing-the-sonar-pro-api/}},
}

@article{alzubi2025open,
    title={Open deep search: Democratizing search with open-source reasoning agents},
    author={Alzubi, Salaheddin and Brooks, Creston and Chiniya, Purva and Contente, Edoardo and von Gerlach, Chiara and Irwin, Lucas and Jiang, Yihan and Kaz, Arda and Nguyen, Windsor and Oh, Sewoong and others},
    journal={arXiv preprint arXiv:2503.20201},
    url={https://arxiv.org/abs/2503.20201},
    year={2025}
}

@misc{grok3_beta,
    author       = {x{AI}},
    title        = {Grok 3 Beta — The Age of Reasoning Agents},
    year         = {2025},
    howpublished = {\url{https://x.ai/news/grok-3/}},
}

@misc{simpleqa_leaderboard,
    author       = {Lopez, Elijah},
    title        = {{AI} {S}imple{QA} Leaderboard},
    year         = {2025},
    howpublished = {\url{https://blog.elijahlopez.ca/posts/ai-simpleqa-leaderboard/}},
}

@misc{introducing_gemini25_pro,
    author       = {Google},
    title        = {Introducing Gemini 2.5 Pro},
    year         = {2025},
    howpublished = {\url{https://blog.google/innovation-and-ai/models-and-research/google-deepmind/gemini-model-thinking-updates-march-2025/#gemini-2-5-pro/}},
}

@misc{introducing_gpt45,
    author       = {Open{AI}},
    title        = {Introducing GPT-4.5},
    year         = {2025},
    howpublished = {\url{https://openai.com/index/introducing-gpt-4-5/}},
}

@article{guo2025deepseek,
    title={Deepseek-r1: Incentivizing reasoning capability in llms via reinforcement learning},
    author={Guo, Daya and Yang, Dejian and Zhang, Haowei and Song, Junxiao and Zhang, Ruoyu and Xu, Runxin and Zhu, Qihao and Ma, Shirong and Wang, Peiyi and Bi, Xiao and others},
    journal={arXiv preprint arXiv:2501.12948},
    url={https://arxiv.org/abs/2501.12948},
    year={2025}
}

@misc{exa_2025,
    author       = {Bryk, Will},
    title        = {State-of-the-art web search API},
    year         = {2025},
    howpublished = {\url{https://exa.ai/blog/api-evals/}},
}

@misc{linkup_2025,
    author       = {Mizrahi, Philippe},
    title        = {Linkup establishes SOTA performance on SimpleQA},
    year         = {2025},
    howpublished = {\url{https://www.linkup.so/blog/linkup-establishes-sota-performance-on-simpleqa/}},
}

@misc{liner_pro_reasoning,
    author       = {Liner},
    title        = {},
    year         = {2025},
    howpublished = {\url{https://liner.com/}},
}

@misc{introducing_gpt41,
    author       = {Open{AI}},
    title        = {Introducing GPT-4.1},
    year         = {2025},
    howpublished = {\url{https://openai.com/index/gpt-4-1/}},
}

@misc{hello_gpt4o,
    author       = {Open{AI}},
    title        = {Hello GPT-4o},
    year         = {2024},
    howpublished = {\url{https://openai.com/index/hello-gpt-4o/}},
}

@misc{introducing_claude4,
    author       = {Anthropic},
    title        = {Introducing Claude 4},
    year         = {2025},
    howpublished = {\url{https://www.anthropic.com/news/claude-4/}},
}

@inproceedings{yang-etal-2018-hotpotqa,
    title = "{H}otpot{QA}: A Dataset for Diverse, Explainable Multi-hop Question Answering",
    author = "Yang, Zhilin  and
      Qi, Peng  and
      Zhang, Saizheng  and
      Bengio, Yoshua  and
      Cohen, William  and
      Salakhutdinov, Ruslan  and
      Manning, Christopher D.",
    editor = "Riloff, Ellen  and
      Chiang, David  and
      Hockenmaier, Julia  and
      Tsujii, Jun{'}ichi",
    booktitle = "Proceedings of the 2018 Conference on Empirical Methods in Natural Language Processing",
    month = oct # "-" # nov,
    year = "2018",
    address = "Brussels, Belgium",
    publisher = "Association for Computational Linguistics",
    url = "https://aclanthology.org/D18-1259/",
    doi = "10.18653/v1/D18-1259",
    pages = "2369--2380",
}

@inproceedings{gong2025papillon,
    title={$\{$PAPILLON$\}$: Efficient and stealthy fuzz $\{$Testing-Powered$\}$ jailbreaks for $\{$LLMs$\}$},
    author={Gong, Xueluan and Li, Mingzhe and Zhang, Yilin and Ran, Fengyuan and Chen, Chen and Chen, Yanjiao and Wang, Qian and Lam, Kwok-Yan},
    booktitle={34th USENIX Security Symposium (USENIX Security 25)},
    pages={2401--2420},
    year={2025}
}
